\documentclass[a4paper]{article}

\usepackage{textcomp}
\usepackage[utf8]{inputenc}
\usepackage[T1]{fontenc}
\usepackage[round]{natbib}
\usepackage[hyphens]{url}
\usepackage{hyperref}
\usepackage[dvipsnames]{xcolor}
\usepackage{amsfonts}
\usepackage{amssymb}
\usepackage{amsmath}
\usepackage{amsthm}
\usepackage{mathtools}
\usepackage{booktabs}
\usepackage{nicefrac}
\usepackage{multirow}
\usepackage{graphicx}
\usepackage{microtype}
\usepackage{array}
\usepackage{caption}
\usepackage{float}
\usepackage[section]{placeins}
\usepackage{tabularx}
\usepackage{threeparttable}
\usepackage{makecell}
\usepackage{listings}

\theoremstyle{plain}

\theoremstyle{definition}

\theoremstyle{remark}

\lstdefinestyle{feedbackprompt}{
    basicstyle=\ttfamily\footnotesize,
    backgroundcolor=\color{black!3},
    breaklines=true,
    columns=fullflexible,
    frame=single,
    framerule=0.4pt,
    framesep=6pt,
    keepspaces=true,
    rulecolor=\color{black!35},
    xleftmargin=0.04\textwidth,
    xrightmargin=0.04\textwidth,
}

\newcommand{\workflowicon}[1]{\raisebox{-0.5\height}{\includegraphics[height=1.75em]{figures/#1.pdf}}}
\newcommand{\smallerworkflowicon}[1]{\raisebox{-0.5\height}{\includegraphics[height=1.55em]{figures/#1.pdf}}}
\newcommand{\largerworkflowicon}[1]{\raisebox{-0.5\height}{\includegraphics[height=2.25em]{figures/#1.pdf}}}

\begin{document}

\begin{center}
{\Large When Does LLM Orchestration Pay Off? A Controlled Evaluation of Accuracy, Cost, and Task Difficulty}
\end{center}

\vspace{7mm}

\noindent\textbf{Nicolas Leins}\hfill\href{mailto:leins@zib.de}{\ttfamily leins@zib.de}\\
\emph{\small Zuse Institute Berlin \& TU Berlin, Berlin, Germany}\\
\\
\textbf{Nico Pelleriti}\hfill\href{mailto:pelleriti@zib.de}{\ttfamily pelleriti@zib.de}\\
\emph{\small Zuse Institute Berlin \& TU Berlin, Berlin, Germany}\\
\\
\textbf{Jana Gonnermann-Müller}\hfill\href{mailto:gonnermann-mueller@zib.de}{\ttfamily gonnermann-mueller@zib.de}\\
\emph{\small Zuse Institute Berlin \& Weizenbaum Institute Berlin, Berlin, Germany}\\
\\
\textbf{Sebastian Pokutta}\hfill\href{mailto:pokutta@zib.de}{\ttfamily pokutta@zib.de}\\
\emph{\small Zuse Institute Berlin \& TU Berlin, Berlin, Germany}\\

\vspace{5mm}

\begin{center}
\begin{minipage}{0.85\textwidth}
\begin{center}
\textbf{Abstract}
\end{center}
{\small
LLM orchestration is often assumed to improve reasoning by allocating additional inference-time computation, yet its gains may not justify its cost. 
Existing comparisons also frequently overlook differences in optimization effort, making it difficult to isolate the value of orchestration itself. 
We conduct a controlled evaluation of Self-Refine, Best-of-$N$, and Debate against task-only and chain-of-thought (CoT) single-call baselines across five LLM backbones and three domains: competitive programming, chess puzzles, and mathematics. 
For comparability, we optimize each method with GEPA under the same optimization budget and evaluate all methods on the same difficulty-stratified benchmark items. 
Orchestration yields moderate but benchmark-dependent gains: averaged across backbones within each benchmark, the largest improvement is 4.6 percentage points over optimized CoT inference and 4.5 points over task-only inference, while requiring approximately 2 to 4 times the mean total tokens of task-only inference. 
Human-derived difficulty is associated with lower absolute accuracy in all three benchmarks, but within-benchmark analyses do not indicate that orchestration effects increase with task difficulty.
By contrast, exploratory mixed-effects analyses reveal strong interactions between orchestration method and backbone model across all three benchmarks, showing that orchestration effectiveness depends substantially on the underlying model. 
Our results suggest that orchestration decisions should be model-specific and account for whether moderate accuracy gains justify the additional inference cost. 
More broadly, evaluations of LLM orchestrations should control optimization effort and report model-specific accuracy–cost trade-offs rather than treating additional inference-time structure as uniformly beneficial.
}
\end{minipage}
\end{center}

\section{Introduction}
\label{intro}

LLMs have become increasingly capable on reasoning-intensive tasks, including mathematical problem solving and code generation~\citep{balunovic_matharena_2025, zhuo_bigcodebench_2025,zimmer_agentic_2026}.
Their performance depends not only on their pretrained weights, but also on how computation is allocated at inference time~\citep{zhou_reasoning_2026}.
One approach to improve performance is to increase test-time computation, for example by generating multiple candidate solutions, critiquing and revising an initial response, or allowing multiple model instances to interact~\citep{snell_scaling_2025,madaan_self-refine_2023,du_improving_2024}.
These methods can improve a model's answer without updating its weights.

In practice, however, performance is not the only objective.
Additional test-time computation consumes more calls and tokens, increases latency, and can introduce further opportunities for failure~\citep{du_survey_2026,cemri_why_2025}.
The central question is whether accuracy gains justify these additional resources.
Evidence is mixed: multi-agent debate and intrinsic self-correction do not reliably outperform simpler alternatives once protocol configuration and inference resources are considered~\citep{smit_should_2024,huang_large_2023}.
Meaningful evaluation therefore requires measuring resource use alongside correctness.

Controlled, within-study evidence about the accuracy--resource trade-offs of common orchestrations remains limited~\citep{du_survey_2026,fang_comprehensive_2025,bai_mas-promptbench_2026}.
Existing comparisons often vary backbones, prompts, stopping criteria, benchmarks, and optimization effort together with workflow structure, making it difficult to attribute observed differences to the evaluated workflows.
Accuracy-only comparisons may also favor complex workflows whose additional components receive more optimization during development~\citep{bai_mas-promptbench_2026}.

Existing difficulty-aware systems use estimated task complexity to select model capacity, allocate test-time computation, or construct query-specific workflows~\citep{bae_complexitynet_2023,cheng_adaptivellm_2025,snell_scaling_2025,su_difficulty-aware_2026,zhang_multi-agent_2025}.
However, these results do not establish that the relative benefit of a fixed orchestration increases monotonically with human-derived task difficulty.
We therefore ask whether fixed orchestrations improve accuracy under a common optimization protocol, what additional inference resources they require, and whether their relative benefits vary systematically with item difficulty.

\subsection{Our Contributions}
\label{sec:contributions}

We address these questions through a controlled, difficulty-stratified evaluation of Self-Refine, Best-of-$N$ (BoN), and Debate against task-only and chain-of-thought (CoT) single-call baselines.

Our main contributions are:
\begin{itemize}
\item We conduct a fully paired comparison of three LLM orchestrations and two single-call baselines across five LLMs and three benchmark domains with item-level difficulty estimates.
We hold backbones, decoding settings, benchmarks, and grading rules fixed, and optimize each orchestration with GEPA under a common maximum optimization budget.
\item We measure the accuracy and resource use of methods and find moderate, benchmark-dependent accuracy gains from orchestration at substantially higher token use.
\item We distinguish the prognostic value of difficulty for predicting absolute accuracy from its potential value for selecting an orchestration.
Although higher human-derived difficulty is associated with lower accuracy, we find no evidence that the relative benefit of the evaluated orchestrations increases systematically with difficulty.
Instead, we find substantial method-by-backbone heterogeneity among the evaluated models.
\end{itemize}

\section{Related Work}
\label{related-work}

We relate our study to research on inference-time orchestration, controlled evaluation of compute and optimization, and difficulty-aware resource allocation.

\subsection{Inference-time LLM orchestration}

CoT prompting elicits intermediate reasoning within a single generation, and shows that a task-agnostic reasoning instruction can provide substantial gains without demonstrations~\citep{wei_chain--thought_2022,kojima_large_2022}.
CoT therefore serves as a single-call reference for multi-call methods.
Multi-call orchestrations allocate additional inference computation through parallel generation, iterative revision, or interaction among model instances.

BoN generates multiple candidate solutions and uses a scorer, verifier, or ranking model to select one.
Independent sampling and voting can improve accuracy as the number of responses increases, although such gains consume additional inference resources and do not isolate the value of coordination~\citep{li_more_2024}.
Candidate diversity and selection can improve over greedy generation, although the gains depend on the available compute and problem difficulty~\citep{snell_scaling_2025}.
Self-Refine orchestrations iteratively generate feedback on an initial response and revise the response using that feedback.
This procedure has improved over one-step generation across diverse domains~\citep{madaan_self-refine_2023}.
Debate orchestrations use multiple model instances to propose solutions, exchange critiques, and produce or select a final answer.
Prior evaluations report improvements in mathematical and strategic reasoning and factuality~\citep{du_improving_2024}.

However, additional reasoning through orchestration does not guarantee higher accuracy.
Long reasoning can produce overthinking, self-correction can overturn correct answers, and Debate may not outperform ensemble-style baselines once accuracy, cost, and latency are considered~\citep{chen_towards_2026,huang_large_2023,smit_should_2024}.

\subsection{Controlling compute, optimization, and workflow failures}

Controlled comparisons address two complementary questions.
A budget-matched design asks which workflow uses a fixed amount of inference computation most effectively, whereas a natural-execution design asks whether each fixed workflow's accuracy justifies the resources it actually consumes.
\citet{tran_single-agent_2026} take the first approach and find that single-agent reasoning matches or outperforms several multi-agent structures under a global thinking-token budget.
We take the second approach by fixing execution rules and measuring the resulting token and call consumption rather than matching inference-token budgets.

Offline optimization effort must also be separated from per-instance inference effort.
Debate rankings can change after protocol tuning, and prompt-optimization effects vary with task and protocol~\citep{smit_should_2024,bai_mas-promptbench_2026}.

Accuracy alone also obscures execution behavior.
OckBench shows that systems with similar reasoning accuracy can differ substantially in token use, establishing efficiency as a distinct outcome~\citep{du_ockbench_2026}.
Listed API prices can also misrepresent realized per-query cost because reasoning-token consumption and interaction counts vary across models~\citep{chen_price_2026}.
MAST attributes failures in multi-agent frameworks to system design, inter-agent misalignment, and task verification, indicating that workflow reliability is not reducible to backbone accuracy~\citep{cemri_why_2025}.
Complete-workflow measurement is therefore needed to expose workflow-level inefficiency and failures.

\subsection{Difficulty measurement, adaptation, and backbone dependence}

Difficulty-aware systems report improvements from adapting inference decisions to estimated task complexity.
ComplexityNet and AdaptiveLLM use complexity estimates for model selection, compute-optimal scaling varies the form and amount of test-time computation, and DAAO and MaAS construct query-specific workflows~\citep{bae_complexitynet_2023,cheng_adaptivellm_2025,snell_scaling_2025,su_difficulty-aware_2026,zhang_multi-agent_2025}.
However, these results do not establish a general relationship between task difficulty and the relative benefit of orchestration.
Because these systems jointly adapt models, resources, or workflow components, their gains may reflect the learned allocation policy rather than difficulty itself.

Beyond item difficulty, the underlying backbone may provide another source of variation in orchestration gains.
Effective test-time scaling strategies vary with base-model capability, and debate hyperparameters transfer only partially across GPT-3.5, GPT-4, and Mixtral~\citep{snell_scaling_2025,smit_should_2024}.
Controlled evaluations further identify single-agent baseline capability as a strong predictor of whether multi-agent coordination helps, with benefits diminishing beyond an empirical capability-saturation threshold~\citep{kim_capable_2026}.
Mechanistically, Self-Refine depends on self-diagnosis, BoN on diversity and selection, and Debate on critique and synthesis; these component capabilities may differ across backbones.
Automated methods such as AFlow, MaAS, and DAAO instead search for or select workflows~\citep{zhang_multi-agent_2025,su_difficulty-aware_2026,zhang_aflow_2025}.
Prior work, therefore, provides limited controlled evidence on how fixed workflows trade accuracy for measured resource use across task difficulty and backbones.

\section{Experimental Design}
\label{sec:experimental-design}

Our experimental design evaluates whether LLM orchestration improves accuracy enough to justify its additional resource use. 
It also tests whether orchestration effects vary with item difficulty or backbone model. 
The following subsections describe the benchmarks, models, workflows, optimization procedure, outcome measures, and statistical analyses.

\subsection{Benchmarks}

We select benchmarks according to three requirements: item-level difficulty estimates, dedicated training data for prompt optimization, and automatic outcome verification.
Easy2Hard-Bench is well suited to these requirements because it augments tasks from multiple domains with item-level difficulty estimates~\citep{ding_easy2hard-bench_2024}.
From this suite, we select Lichess (chess puzzles), Codeforces (programming), and AMC (mathematics).
All provide continuous, human-derived item-difficulty estimates: Lichess and Codeforces use ratings derived from player records, whereas AMC uses Item Response Theory (IRT) fit to human item statistics.
We use dedicated training splits for GEPA and fixed 200-item subsets of the disjoint evaluation splits.
The subsets make the fully crossed model--method evaluation computationally feasible while retaining adequate statistical power.
For each E2H benchmark, we stratify items jointly by decile of the normalized difficulty quantile and tercile of rating uncertainty, and allocate the evaluation sample proportionally across strata.
This preserves the source difficulty distribution; sharing each frozen subset across all models and methods prevents subset selection from confounding comparisons.

\subsection{Models}

We evaluate five locally hosted, open-weight LLMs: DeepSeek V4 Flash, Gemma 4, GLM 4.7 Flash, Qwen 3.6, and GPT-OSS Puzzle 88B.
Within each experiment, all methods share a backbone and decoding settings.
All models use the same 120{,}000-token output limit to limit inference-time computation.
Models are served with vLLM using the recommended inference settings (temperature 1.0 and top-$p$ 0.95).
Qwen3.6-35B-A3B-FP8 runs on two NVIDIA A100 80\,GB GPUs; GLM-4.7-Flash, Gemma 4 26B-A4B-it, and NVIDIA GPT-OSS Puzzle 88B each run on one NVIDIA H200 141\,GB GPU; and DeepSeek-V4-Flash runs on four H200 GPUs with reasoning effort set to \texttt{high}.
For particularly long GEPA runs with DeepSeek as both task and profile model, we also use the official DeepSeek API with the same inference settings.

\subsection{Orchestrations}
\label{sec:orchestrations}
We evaluate task-only and GEPA-optimized CoT single-call baselines together with Self-Refine, BoN, and Debate.
We use ``orchestration'' to refer to the three multi-call methods in this comparison.
Table~\ref{tab:orchestrations} summarizes the compared orchestrations and their control structures.

\begin{table}[t]
\centering
\setlength{\tabcolsep}{2pt}
\renewcommand{\arraystretch}{1.2}
\begin{threeparttable}
\caption{Compared orchestration variants and their high-level control structure.}
\label{tab:orchestrations}
\begin{tabular}{@{}l c l c@{}}
\toprule
\textbf{method} & \textbf{workflow} & \textbf{control structure} & \textbf{calls} \\
\midrule
LLM & \smallerworkflowicon{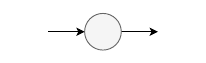} & task prompt only & 1 \\
CoT LLM & \smallerworkflowicon{singleLLM} & task + CoT system prompt & 1 \\
\addlinespace
Self-Refine & \largerworkflowicon{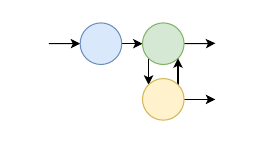} & generate $\rightarrow$ feedback $\rightarrow$ refine & $2\leq 5$ \\
BoN & \largerworkflowicon{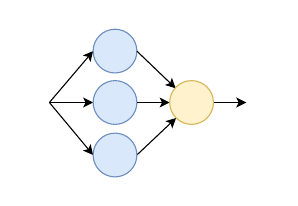} & 3 independent samples + selection & 4 \\
Debate & \workflowicon{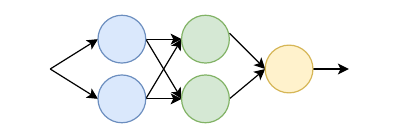} & 2 samples $\rightarrow$ 1 debate round $\rightarrow$ judge synthesis & 5 \\
\bottomrule
\end{tabular}
\end{threeparttable}
\end{table}

The LLM baseline uses the benchmark task prompt without an additional system prompt.
CoT LLM adds a GEPA-optimized CoT system prompt.
Self-Refine produces an initial solution followed by at most two feedback--refinement rounds~\citep{madaan_self-refine_2023}.
It stops when the feedback states ''no further refinement is needed''; otherwise, it completes both rounds.
BoN selects among three independent candidates by parsing the final recognized selector statement.
In Debate, two agents independently answer, revise after observing each other, and pass both revisions to a judge.
It always executes all five calls without consensus- or confidence-based early stopping.

\subsection{Budget-controlled prompt optimization with GEPA}
\label{sec:gepa}

An orchestration's control structure defines which model calls are made and how information flows between them, but this structure is only a scaffold.
Each call must also be instructed about its role, the intermediate information it should evaluate, and the output it should pass to subsequent calls.
Making a scaffold effective can therefore require substantial prompt engineering, particularly for workflows containing several specialized calls.

Manual prompt engineering introduces degrees of freedom that are difficult to standardize.
Some methods may receive more iteration, and researchers may incorporate different amounts of task knowledge or stop tuning methods at different stages.
A comparison of hand-crafted prompts can consequently conflate the effectiveness of an orchestration with the effort invested in configuring it.

We address this confound by treating prompt development as a budgeted optimization problem.
After fixing each method's scaffold, we use GEPA to optimize its textual prompt components~\citep{agrawal_gepa_2026}.
GEPA automates candidate generation, evaluation, and refinement under common metric-call and weighted-token limits.
The following abridged Codeforces CoT example illustrates how GEPA refined the seed prompt:

\begin{lstlisting}[style=feedbackprompt]
Think step by step through the competitive programming problem before giving your final answer.
Follow the required output format.
\end{lstlisting}

The final selected candidate prompt was:

\begin{lstlisting}[style=feedbackprompt]
Before writing code, analyze the problem, identify the algorithm, and consider edge cases.
[detailed line-by-line input requirements]
Handle edge cases: empty input, single elements, duplicates, boundaries.
[output-contract instructions]
Follow these guidelines to minimize runtime errors and ensure correct output.
\end{lstlisting}

The simple seed prompt asks the model to reason step by step and follow the task's output format, but leaves the intended procedure largely unspecified.
The selected candidate operationalizes this instruction by separating algorithm design from implementation, encoding the execution environment's input constraint, and imposing an explicit output contract.
The corresponding validation accuracy increased from .20 for the seed to .82 for the selected candidate.

Each benchmark contributes a 100-item subset of its training split for GEPA optimization.
We split it into 50 training and 50 validation items using the same subsetting procedure described above.
GEPA jointly optimizes all designated prompt components of each method.
For CoT, it optimizes the generation system prompt.
For Self-Refine, it optimizes the generation system prompt, feedback template, and refinement template.
For BoN, it optimizes the generation system prompt, selector system prompt, and selector task template.
For Debate, it optimizes the generation system prompt, debate template, and judge template.
For every optimized method, GEPA receives a maximum of 750 metric calls, following the recommended 15 calls per training example.
We additionally cap weighted token consumption at the cost of a full CoT GEPA run: 40M for Codeforces and 75M for Lichess.
Together, these limits give every method within a benchmark the same maximum optimization budget.
Weighted tokens $t_w$ are defined as $\text{input tokens} + w \times (\text{output} + \text{reasoning tokens})$ with $w=4$ representing an estimate of the average cost ratio of input to output tokens of common providers.
The design therefore compares optimization under a common resource constraint rather than equalizing iterations.
We perform one GEPA run for each method--benchmark combination, using DeepSeek V4 Flash as the task, profile, and reflection model.
We select the candidate with the highest validation accuracy, breaking ties in favor of the earliest candidate.
For AMC, GEPA did not improve on the near-saturated seed configuration, so we instead use a hand-crafted prompt that modestly improves on the seed.
This manual intervention is an exception to the otherwise automated prompt-refinement protocol.
It avoids placing AMC at a prompt-quality disadvantage relative to benchmarks where GEPA improved the configuration.

\subsection{Grading and failure handling}
All three benchmarks are graded as pass@1: each instance receives a single attempt that is scored as correct (1) or incorrect (0) under benchmark-specific rules.
For Lichess and AMC, we parse the output and require an exact reference match; for Codeforces, the generated code must pass all provided tests from the benchmark.
We retry transient technical and malformed-generation failures, but retain maximum-output-limit failures as score-zero outcomes.
BoN and Debate continue after an empty intermediate response because later independent calls and selection or synthesis may still produce a valid answer.
We estimate uncertainty for orchestration contrasts with 9,999 stratified whole-item bootstrap replicates that preserve the paired method and model observations for each sampled item.
Descriptive model-efficiency intervals use 9,999 unstratified whole-item bootstrap replicates, while single-call baseline comparisons use paired normal-approximation intervals.

\subsection{Measurements}
For each task attempt, we log correctness, input tokens, output tokens, and reasoning tokens.
Because monetary cost and latency vary by provider and serving configuration, we use token consumption as a reproducible proxy for inference resource use~\citep{du_ockbench_2026,huan_scaling_2025}. But to account for higher output costs than input costs, we use the weighted token $t_w$ metric described above.
The primary effectiveness metric is accuracy, the fraction of tasks solved correctly.
For orchestration $o$ evaluated on task set $\mathcal{T}$, we compute $\mathrm{Acc}(o)=\frac{\#\text{ correct}}{\#\mathcal{T}}$.
To relate effectiveness to cost, we report token efficiency as $\mathrm{TokenEff}(o)=\frac{\overline{t_w}(o)}{\mathrm{Acc}(o)}$.
We report all metrics overall and stratified by within-benchmark difficulty quartile.

\subsection{Statistical analysis}
Raw accuracy differences do not indicate whether an observed gap is systematic or could reflect finite item sampling, shared item difficulty, or repeated evaluation of the same item across methods.
We therefore complement descriptive results with a structured inferential analysis that preserves pairing, quantifies uncertainty, accounts for item-level dependence, and corrects planned families of comparisons for multiplicity.
This design provides a more stringent assessment of orchestration effects than unadjusted comparisons of aggregate benchmark scores.

As a descriptive validity check, we compute Spearman correlations between continuous difficulty and item-level accuracy separately for each benchmark, averaging accuracy over the balanced five-method by five-LLM design.
We obtain 95\% percentile intervals from 9,999 whole-item bootstrap replicates and apply Holm correction across the three benchmark-level tests.

We fit mixed models separately by benchmark to the complete paired observations for CoT, Self-Refine, BoN, and Debate.
We prespecified CoT as the reference because we expected it to be stronger than task-only inference.
Task-only is therefore excluded from the mixed models but retained in descriptive comparisons.
Let $Y=\texttt{correct}\in\{0,1\}$ denote per-attempt correctness.
All models use a binomial family with a logit link, the fixed effects specified below, and random intercepts for item and item-by-LLM pairs.
These random intercepts account for residual item difficulty and dependence among methods evaluated on the same item--LLM pair.
We standardize continuous difficulty separately within each benchmark.

For each benchmark, we first fit Model 0 ($M_0$) and Model 1 ($M_1$):

\medskip
{\small\noindent\(
\begin{aligned}
M_0:\quad &\texttt{correct} \sim \texttt{difficulty} + \texttt{LLM} + (1 \mid \texttt{item}) + (1 \mid \texttt{item}:\texttt{LLM}), \\
M_1:\quad &\texttt{correct} \sim \texttt{method} + \texttt{difficulty} + \texttt{LLM} + (1 \mid \texttt{item}) + (1 \mid \texttt{item}:\texttt{LLM}).
\end{aligned}
\)\par}
\medskip

We compare $M_0$ and $M_1$ using a likelihood-ratio test to assess whether the method affects accuracy on average, after adjusting for difficulty and the five fixed LLMs.
The three planned $M_1$ contrasts compare Self-Refine, BoN, and Debate with CoT.
$M_2$ adds the method-by-difficulty interaction.

\medskip
{\small\noindent\(
\begin{aligned}
M_2:\quad &\texttt{correct} \sim \texttt{method} * \texttt{difficulty} + \texttt{LLM} + (1 \mid \texttt{item}) + (1 \mid \texttt{item}:\texttt{LLM}).
\end{aligned}
\)\par}
\medskip

We compare $M_1$ and $M_2$ using a likelihood ratio test to assess whether allowing method-specific difficulty slopes improves model fit.
We apply Holm correction across these three contrasts within each benchmark.
Difficulty quartiles are used only to preserve the stratified item-sampling design in bootstrap resampling and are not included as inferential predictors.

Based on the descriptive results, we additionally fit exploratory $M_3$, which retains $M_1$'s common difficulty slope and adds a method-by-LLM interaction.

\medskip
{\small\noindent\(
\begin{aligned}
M_3:\quad &\texttt{correct} \sim \texttt{method} * \texttt{LLM} + \texttt{difficulty} + (1 \mid \texttt{item}) + (1 \mid \texttt{item}:\texttt{LLM}).
\end{aligned}
\)\par}
\medskip

Its likelihood-ratio comparison with $M_1$ tests whether average method effects vary across the five evaluated LLMs.

\section{Results}
\label{results}

\subsection{GEPA optimization and single-call baselines}
On Codeforces, the best-candidate accuracies generated by GEPA on the validation set were 82\% (+62 points over the seed) for CoT, 76\% (+28) for Self-Refine, 74\% (+22) for BoN, and 80\% (+0) for Debate.
On Lichess, they were 56\% (+14) for CoT, 54\% (+4) for Self-Refine, 52\% (+0) for BoN, and 54\% (+6) for Debate.
Supplementary Material provides additional details on the GEPA runs and reports full best-candidate prompts.

We first assess whether the CoT prompt improves the single-call task-only baseline.
Across the five models, CoT differed from task-only by $-0.1$ percentage points on Codeforces (paired normal-approximation 95\% CI $[-2.8, 2.6]$), $+1.5$ points on Lichess (95\% CI $[-1.0, 4.0]$), and $-1.4$ points on AMC (95\% CI $[-2.9, 0.1]$).
These benchmark-level differences are marginal, and their intervals include zero.
We retain both baselines in the descriptive results.
Despite this, CoT remains the primary inferential reference because it represents the hypothesized stronger single-call reasoning baseline.
The model-specific CoT-minus-task-only differences ranged from $-10.5$ points (GLM) to $+10.0$ (DeepSeek) on Codeforces, from $-3.5$ (GPT-OSS) to $+8.5$ (DeepSeek) on Lichess, and from $-9.0$ (Gemma) to $+3.0$ (Qwen) on AMC.

\subsection{Average effects of orchestration}
\begin{figure}[t]
\centering
\includegraphics[width=0.75\textwidth]{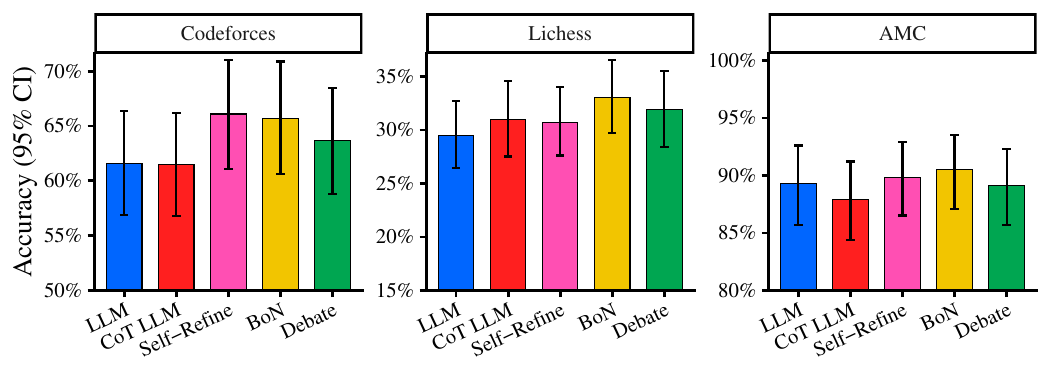}
\caption{Accuracy by method and benchmark, aggregated over the five evaluated models.
Bars are point estimates, and error bars show 95\% confidence intervals.}
\label{fig:overall-accuracy}
\end{figure}

We next test whether orchestration method affects accuracy on average across the five evaluated LLMs.
Adding method improved mixed-model fit for Codeforces ($\chi^2(3)=18.62$, $p=.0003$) and AMC ($\chi^2(3)=13.76$, $p=.0032$), but not Lichess ($\chi^2(3)=4.34$, $p=.227$).
The corresponding AIC comparisons favored $M_1$ for Codeforces and AMC and $M_0$ for Lichess.
The overall accuracies of task-only, CoT, Self-Refine, BoN, and Debate were 61.6\%, 61.5\%, 66.1\%, 65.7\%, and 63.7\% on Codeforces; 29.5\%, 31.0\%, 30.7\%, 33.0\%, and 31.9\% on Lichess; and 89.3\%, 87.9\%, 89.8\%, 90.5\%, and 89.1\% on AMC (see Figure~\ref{fig:overall-accuracy}).
Relative to task-only inference, the three orchestration differences ranged from $+2.1$ points (Debate) to $+4.5$ (Self-Refine) on Codeforces, from $+1.2$ (Self-Refine) to $+3.5$ (BoN) on Lichess, and from $-0.2$ (Debate) to $+1.2$ (BoN) on AMC.
Relative to CoT, the corresponding paired differences ranged from $+2.2$ points (Debate) to $+4.6$ (Self-Refine) on Codeforces, from $-0.3$ (Self-Refine) to $+2.0$ (BoN) on Lichess, and from $+1.2$ (Debate) to $+2.6$ (BoN) on AMC.
The 95\% whole-item bootstrap intervals excluded zero for Codeforces Self-Refine ($+4.6$ points, $[2.1, 7.1]$) and BoN ($+4.2$, $[1.8, 6.5]$), and for AMC Self-Refine ($+1.9$ points, $[0.5, 3.3]$) and BoN ($+2.6$, $[1.3, 4.0]$); the other CoT contrasts included zero.
Relative to CoT, $M_1$ estimated higher average odds of success for Self-Refine and BoN on Codeforces (OR $=1.90$, 95\% CI $[1.36,2.65]$, Holm-adjusted $p=.0004$; and OR $=1.79$, 95\% CI $[1.29,2.49]$, $p=.0010$) and AMC (OR $=1.99$, 95\% CI $[1.16,3.40]$, $p=.0238$; and OR $=2.68$, 95\% CI $[1.53,4.68]$, $p=.0016$).
No method contrast was significant on Lichess, and Debate did not significantly outperform CoT on any benchmark.

\subsection{Effects across task difficulty}

Difficulty and item-level accuracy were negatively correlated on Codeforces (Spearman's $\rho=-.533$, 95\% CI $[-.641,-.409]$), Lichess ($\rho=-.660$, 95\% CI $[-.734,-.571]$), and AMC ($\rho=-.485$, 95\% CI $[-.591,-.362]$); all Holm-adjusted $p<.001$.
Thus, higher human-derived difficulty was associated with lower aggregate accuracy in all three benchmarks.
The empirical ten-bin accuracy trajectories were not monotonic throughout, however.
For example, task-only accuracy decreased from 92.3\% to 37.9\% between the lowest and highest Codeforces bins and from 67.3\% to 6.2\% on Lichess.
On AMC, the corresponding bin estimates were 99.0\% and 52.0\%, with a local increase to 98.0\% in bin 9.
The corresponding ten-bin trajectories are shown in the Supplementary Material.

\begin{figure}[t]
\centering
\includegraphics[width=0.75\textwidth]{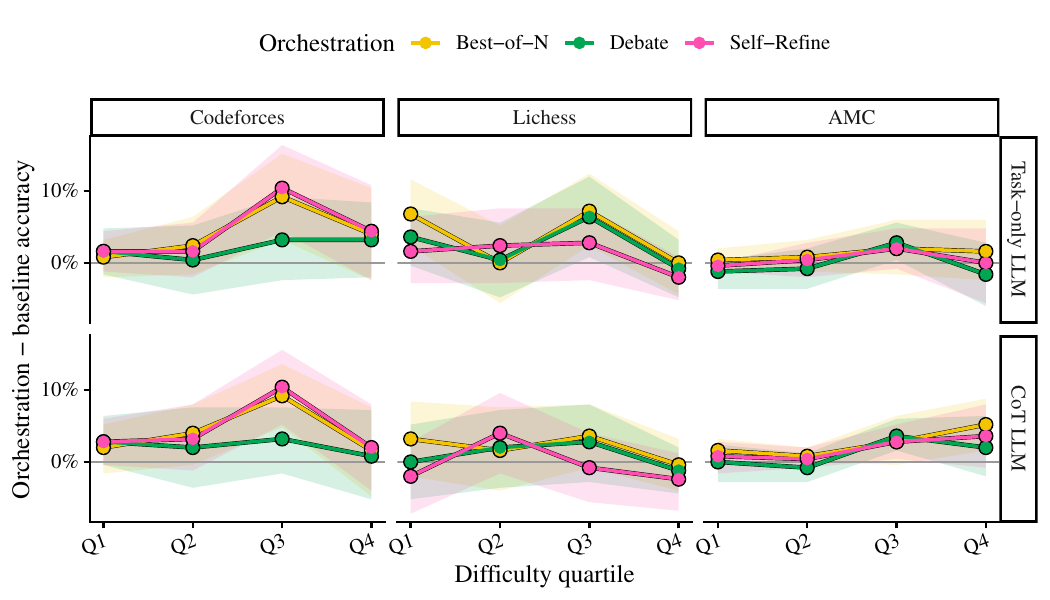}
\caption{Orchestration-minus-baseline accuracy by within-benchmark difficulty quartile.
Differences are aggregated over the five evaluated models and in each quartile.
Panels distinguish benchmarks and task-only and CoT baselines, and confidence bands show 95\% bootstrap CI.}
\label{fig:difficulty-differences}
\end{figure}

We next assessed whether the relative advantage of orchestration changed across difficulty levels.
The orchestration-minus-single-LLM differences varied across descriptive difficulty quartiles without a common monotonic pattern.
Figure~\ref{fig:difficulty-differences} shows these differences relative to both task-only and CoT.
On Codeforces, the largest quartile differences occurred in Q3 for Self-Refine ($+10.4$ points) and BoN ($+9.2$ points), whereas their Q4 differences were $+2.0$ and $+1.6$ points.
On Lichess, Q4 differences ranged from $-2.4$ to $-0.4$ points, while differences across the other quartiles ranged from $-2.0$ to $+4.0$ points.
On AMC, the differences ranged from $-0.8$ to $+5.2$ points, with the largest value for BoN in Q4.

The additive model ($M_1$) estimated a common continuous-difficulty slope across methods, whereas the interaction model ($M_2$) allowed method-specific slopes.
Compared with $M_1$, $M_2$ did not improve model fit for Codeforces ($\chi^2(3)=1.47$, $p=.688$), Lichess ($\chi^2(3)=0.23$, $p=.972$), or AMC ($\chi^2(3)=0.70$, $p=.874$).
$M_1$ also had the lowest AIC across all benchmarks.
Thus, the primary analysis did not support the hypothesis that orchestration becomes systematically more beneficial as item difficulty increases.
The nine planned orchestration-versus-optimized-CoT slope ratios ranged from 0.82 to 1.19; every nominal 95\% interval included the null ratio of one, and every within-benchmark Holm-adjusted $p$-value was at least .887.
The Supplementary Material gives the complete tests and contrasts.

\subsection{Variation across LLMs}
\begin{figure}[t]
\centering
\includegraphics[width=0.75\textwidth]{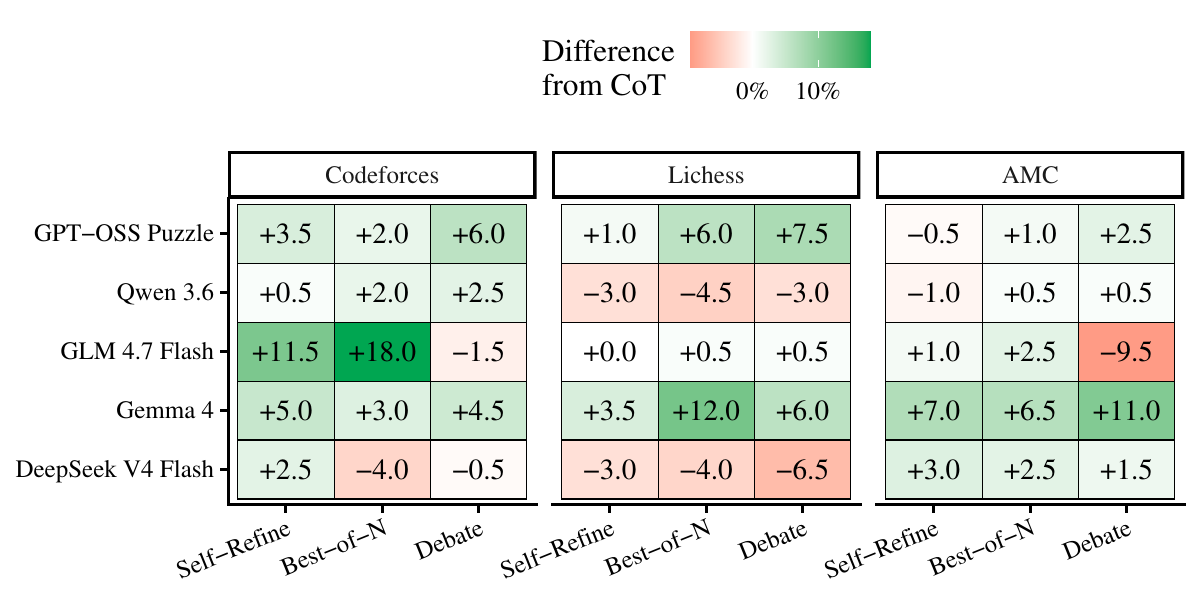}
\caption{Model-specific orchestration-minus-CoT paired accuracy differences by benchmark.
Each cell averages item differences per model; labels are percentage-point differences.}
\label{fig:model-heterogeneity}
\end{figure}

We next assess whether the average method effects were consistent across the five evaluated LLMs.
Exploratory likelihood-ratio tests indicated method-by-LLM heterogeneity on Codeforces ($\chi^2(12)=53.82$, $p<.0001$), Lichess ($\chi^2(12)=30.68$, $p=.0022$), and AMC ($\chi^2(12)=87.68$, $p<.0001$).
The model-specific paired differences from optimized CoT ranged from $-4.0$ points (DeepSeek, BoN) to $+18.0$ (GLM, BoN) on Codeforces, from $-6.5$ (DeepSeek, Debate) to $+12.0$ (Gemma, BoN) on Lichess, and from $-9.5$ (GLM, Debate) to $+11.0$ (Gemma, Debate) on AMC (see Figure~\ref{fig:model-heterogeneity})

\subsection{Inference-resource trade-offs}
\begin{figure}[t]
\centering
\includegraphics[width=0.75\textwidth]{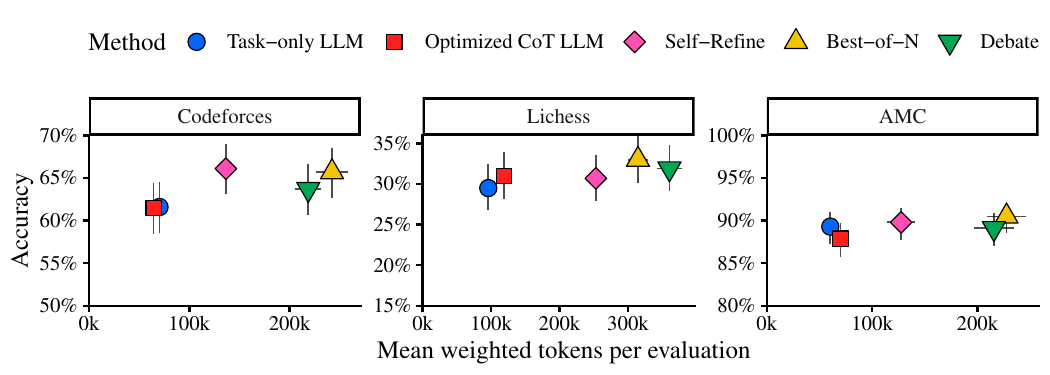}
\caption{Accuracy against mean weighted tokens for each method and benchmark.
Each point aggregates equally over the five evaluated models, and error bars show 95\% confidence intervals on both axes.}
\label{fig:accuracy-resource}
\end{figure}

Mean weighted-token ($t_w$) use for CoT was 0.92 (Codeforces), 1.24 (Lichess), and 1.16 (AMC) times that of task-only inference.
Mean $t_w$ use for Self-Refine, BoN, and Debate was 2.12, 3.76, and 3.40 times CoT on Codeforces; 2.13, 2.65, and 3.04 times on Lichess; and 1.82, 3.25, and 3.08 times on AMC.
In absolute terms, mean $t_w$ per evaluation ranged from 127{,}577 to 227{,}585 for the orchestrations on AMC, from 136{,}487 to 242{,}343 on Codeforces, and from 252{,}993 to 360{,}563 on Lichess.
Figure~\ref{fig:accuracy-resource} shows the resulting relationship between accuracy and mean weighted-token use.

Across model--benchmark cells, task-only accuracy ranged from 5.0\% (GLM, Lichess) to 93.0\% (DeepSeek, AMC), and task-only weighted tokens per correct answer ranged from 26{,}027 (GPT-OSS, AMC) to 2{,}653{,}007 (GLM, Lichess).
The orchestration-to-task-only ratios of weighted tokens per correct answer ranged from 1.41 to 5.15 on Codeforces, from 1.42 to 5.88 on Lichess, and from 1.82 to 4.97 on AMC.
For GPT-OSS on Codeforces, task-only and Debate achieved 58.0\% and 70.0\% accuracy while using 37{,}024 and 130{,}888 weighted tokens per correct answer, respectively.
For Gemma on Lichess, task-only and BoN achieved 33.5\% and 43.0\% accuracy with 467{,}421 and 1{,}015{,}592 weighted tokens per correct answer.
For Gemma on AMC, task-only and Debate achieved 90.5\% and 92.5\% accuracy with 72{,}296 and 359{,}564 weighted tokens per correct answer.
The Supplementary Material reports these and additional model--method cells with descriptive whole-item bootstrap intervals.

Maximum-output-limit outcomes reflect models consuming the available completion budget without returning a valid answer, rather than transient infrastructure failures.
They are therefore evidence of high reasoning consumption and a limitation of the evaluated models under the common output budget.
This pattern was most pronounced for Gemma on Lichess, where 49 of 200 task-only attempts and 97 of 200 optimized-CoT attempts reached the maximum-output limit.
These attempts remain score-zero outcomes, and their token consumption remains included in the resource totals.

\section{Discussion}
\label{discussion}

Our results establish a clear accuracy--resource trade-off.
Orchestration delivered moderate, benchmark-dependent gains while consistently consuming more inference resources, making a validated single-call method the stronger efficiency-oriented default.
Self-Refine and BoN improved over optimized CoT on Codeforces and AMC, but no orchestration significantly improved accuracy on Lichess, and Debate did not significantly outperform CoT on any benchmark.
More calls and richer interaction therefore did not guarantee better outcomes.
Orchestration is justified when its incremental accuracy has sufficient application value to offset the additional token, latency, and serving costs.

On AMC, high single-call accuracy left limited room for improvement. In contrast, performance on the hardest Codeforces and Lichess items suggests a different boundary: additional computation has little value when the backbone cannot generate a viable solution.
This interpretation agrees with prior evidence that test-time computation provides limited gains on the hardest problems~\citep{snell_scaling_2025} and with our Codeforces results, where Self-Refine and BoN achieved their largest gains in the third rather than the fourth difficulty quartile.
Together, these patterns show that neither available headroom nor difficulty alone determines the value of orchestration.

Human-derived difficulty was strongly associated with lower absolute accuracy, yet the relative benefits of Self-Refine, BoN, and Debate did not increase systematically with difficulty.
Identifying tasks that a single call is likely to miss is therefore different from identifying tasks that a particular workflow is likely to correct.
This distinction explains why adaptive systems can benefit from difficulty signals without implying that every hard query should receive more calls.
Such systems can jointly select model capacity, computation, and workflow structure, while learned difficulty can represent model-relative solvability more directly than human assessment~\citep{bae_complexitynet_2023,cheng_adaptivellm_2025,snell_scaling_2025,su_difficulty-aware_2026,zhang_multi-agent_2025}.

Viewed against existing work, these results place orchestration within the broader problem of conditional resource allocation.
Preference-based model routing asks when one model will outperform another, while difficulty-aware systems vary model capacity, computation, or workflow structure across queries~\citep{ong_routellm_2025}.
Our findings show why the success of such policies does not imply a general rule that harder items benefit more from a fixed orchestration.
Instead, the relationship between additional inference and accuracy depends on the workflow and backbone, consistent with prior evidence that effective test-time scaling strategies vary across models~\citep{snell_scaling_2025,smit_should_2024}.

This dependence is also evident in the backbone analysis.
The strong method-by-backbone interactions across all three benchmarks demonstrate that a workflow effective for one model can be neutral or harmful for another.
Revision, selection, critique, and synthesis rely on capabilities that differ across models, so the same workflow structure can produce substantially different outcomes across backbones.
Workflow selection and prompt optimization should therefore be treated as part of model selection, with accuracy and resource use validated for the intended model--workflow pair.

Finally, equalizing the maximum optimization budget strengthens the comparison by separating orchestration structure from unequal tuning effort.
Under this common budget, additional workflow complexity still did not produce uniformly large gains.
Optimization is therefore part of the evaluated method rather than a background implementation detail, and orchestration claims should report both how a workflow was obtained and the inference resources it consumes.

\section{Limitations}
\label{limitations}

Our conclusions are bounded by the evaluated optimization procedure and empirical scope.
The comparison uses GEPA under a common optimization budget, but GEPA may not optimize workflows with different numbers and types of tunable components equally well.
Topology-aware optimizers or different optimization budgets could therefore produce different accuracy--cost trade-offs, particularly for more complex workflows.
Moreover, we evaluate three reasoning domains, five open-weight backbones, and one fixed configuration of each workflow.
This breadth supports comparison across varied settings, while evaluations of additional domains, backbones, and orchestration configurations are needed to establish how broadly the observed trade-offs extend.
Our difficulty analysis concerns a common linear relationship between independently estimated human difficulty and the relative benefit of three fixed workflows.

\section{Conclusion}
\label{conclusion}

Under a common prompt-optimization budget, orchestration delivered moderate, benchmark-dependent accuracy gains at substantially higher inference resource use.
Self-Refine and BoN outperformed optimized CoT on Codeforces and AMC, whereas no orchestration significantly improved over CoT on Lichess, and Debate did not significantly outperform CoT on any benchmark.
Human-derived difficulty tracked absolute accuracy but not the relative benefit of orchestration, while exploratory analyses revealed substantial method-by-backbone heterogeneity.
These findings show that difficulty alone does not determine when orchestration pays off and that evaluations should report workflow- and backbone-specific accuracy--resource trade-offs under controlled optimization.

\bibliographystyle{plainnat}
\bibliography{references}

\clearpage
\appendix
\onecolumn

\begin{center}
{\Large Supplementary Material}
\end{center}

\newcolumntype{L}{>{\raggedright\arraybackslash}p{.20\textwidth}}
\newcommand{\prompttext}[1]{{\tiny\detokenize{#1}\par}}
\section{GEPA Optimization Outcomes and Candidate Trees}
\label{app:gepa-trees}

\begin{figure}[H]
\centering
\begin{minipage}[t]{0.18\textwidth}
\centering
\includegraphics[width=\linewidth,height=.10\textheight,keepaspectratio]{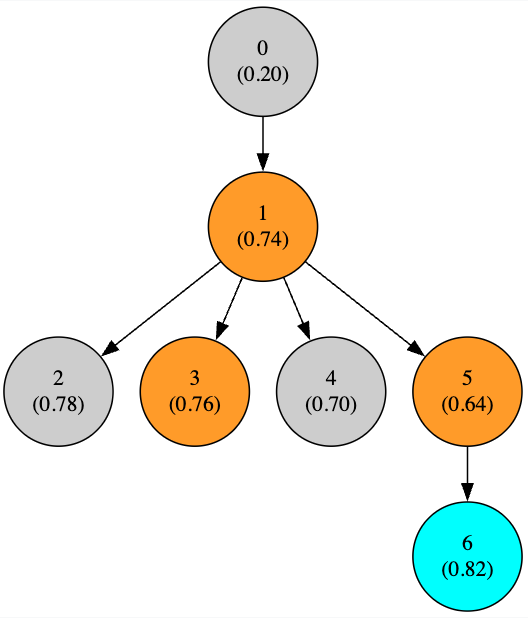}\\
optimized CoT
\end{minipage}
\begin{minipage}[t]{0.18\textwidth}
\centering
\includegraphics[width=\linewidth,height=.10\textheight,keepaspectratio]{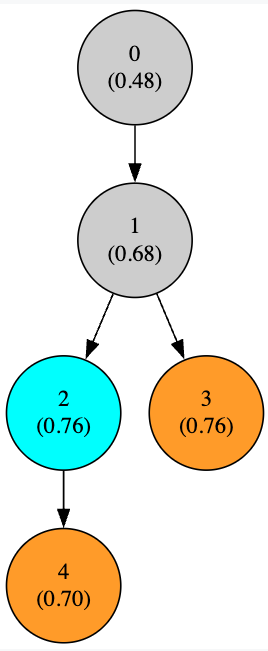}\\
Self-Refine
\end{minipage}
\begin{minipage}[t]{0.18\textwidth}
\centering
\includegraphics[width=\linewidth,height=.10\textheight,keepaspectratio]{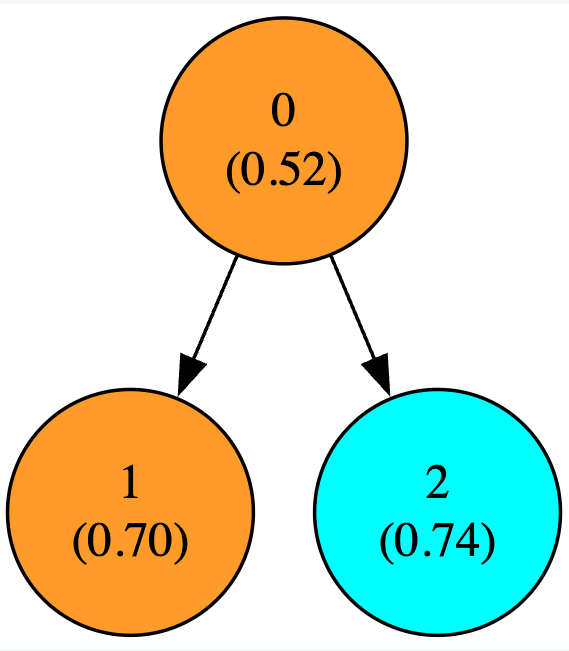}\\
BoN
\end{minipage}
\begin{minipage}[t]{0.18\textwidth}
\centering
\includegraphics[width=\linewidth,height=.10\textheight,keepaspectratio]{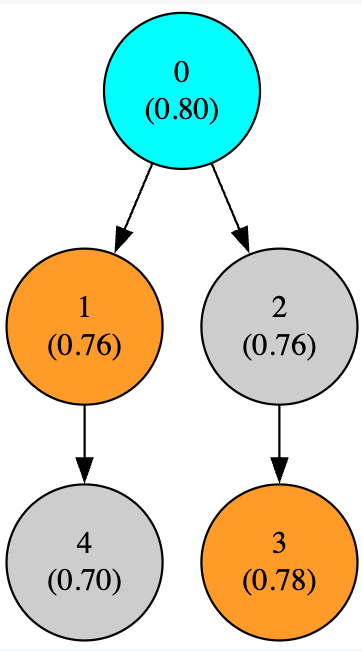}\\
Debate
\end{minipage}
\caption{Archived GEPA candidate trees for Codeforces.
Node labels give candidate identifiers and validation accuracy; cyan marks the selected candidate.}
\label{fig:gepa-codeforces}
\end{figure}

\begin{figure}[H]
\centering
\begin{minipage}[t]{0.18\textwidth}
\centering
\includegraphics[width=\linewidth,height=.10\textheight,keepaspectratio]{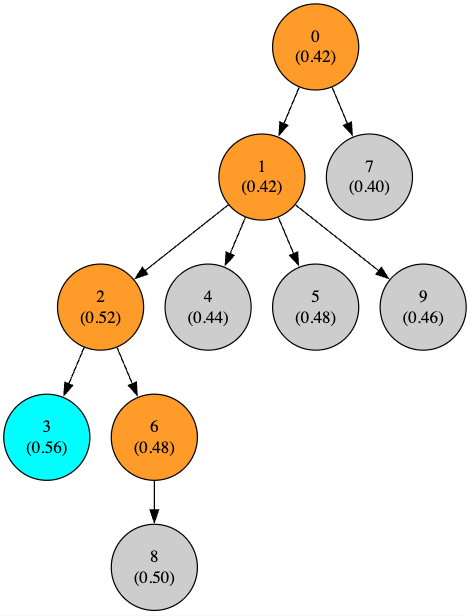}\\
optimized CoT
\end{minipage}
\begin{minipage}[t]{0.18\textwidth}
\centering
\includegraphics[width=\linewidth,height=.10\textheight,keepaspectratio]{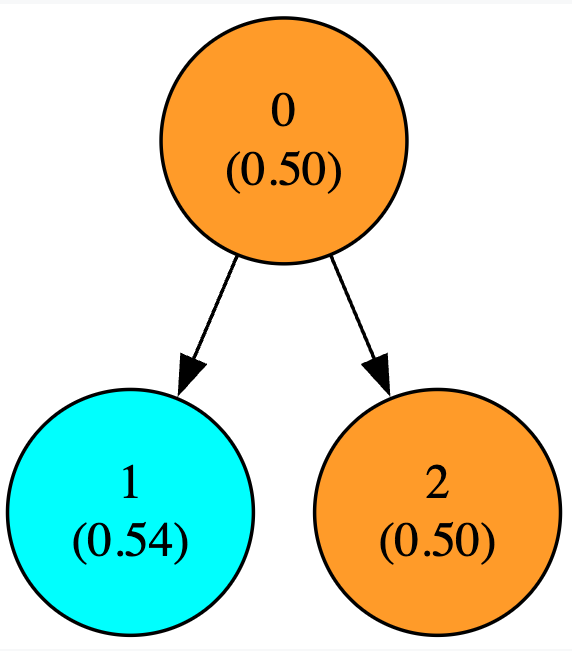}\\
Self-Refine
\end{minipage}
\begin{minipage}[t]{0.18\textwidth}
\centering
\includegraphics[width=\linewidth,height=.10\textheight,keepaspectratio]{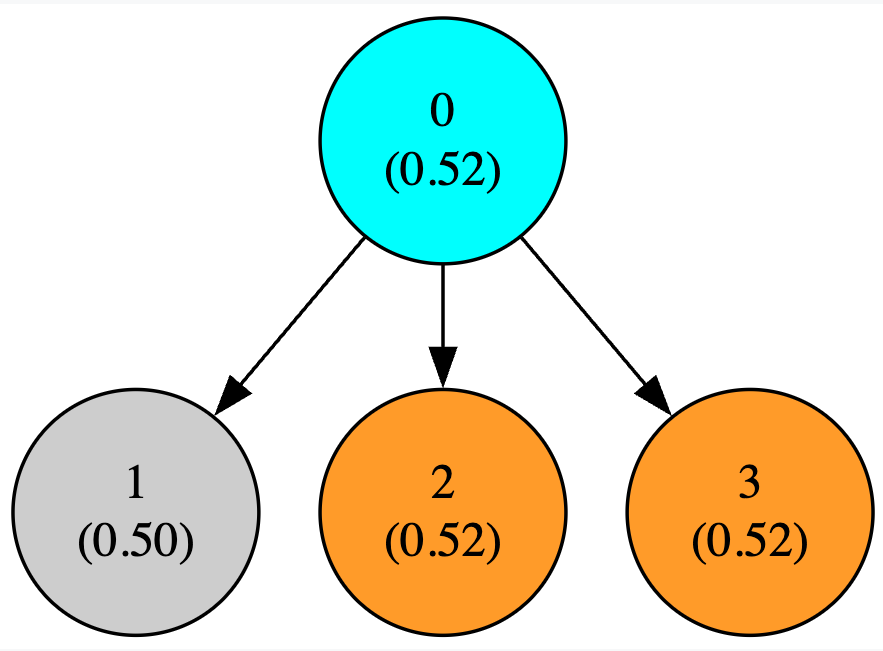}\\
BoN
\end{minipage}
\begin{minipage}[t]{0.18\textwidth}
\centering
\includegraphics[width=\linewidth,height=.10\textheight,keepaspectratio]{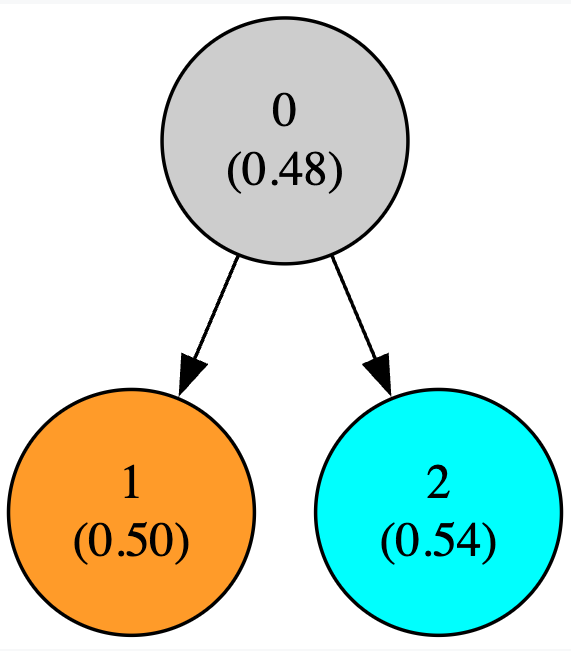}\\
Debate
\end{minipage}
\caption{Archived GEPA candidate trees for Lichess.
Node labels give candidate identifiers and validation accuracy; cyan marks the selected candidate.}
\label{fig:gepa-lichess}
\end{figure}

Table~\ref{tab:gepa-outcomes} summarizes GEPA optimization outcomes for each method on Codeforces and Lichess.
It reports accuracy, added candidates, and metric calls, while Figures~\ref{fig:gepa-codeforces} and~\ref{fig:gepa-lichess} show the corresponding candidate trees.

\begin{table}[H]
\centering
\footnotesize
\caption{GEPA optimization outcomes on the 50-item validation partitions.
Added candidates exclude seed candidate 0.}
\label{tab:gepa-outcomes}
\begin{tabular}{@{}lrrrr@{}}
\toprule
measure & CoT & Self-Refine & BoN & Debate \\
\midrule
\multicolumn{5}{@{}l}{\textit{Codeforces}} \\
seed accuracy & .20 & .48 & .52 & .80 \\
selected accuracy & .82 & .76 & .74 & .80 \\
added candidates & 6 & 4 & 2 & 4 \\
metric calls & 750 & 388 & 216 & 448 \\
\addlinespace[2pt]
\multicolumn{5}{@{}l}{\textit{Lichess}} \\
seed accuracy & .42 & .50 & .52 & .48 \\
selected accuracy & .56 & .54 & .52 & .54 \\
added candidates & 9 & 2 & 3 & 2 \\
metric calls & 750 & 264 & 254 & 204 \\
\bottomrule
\end{tabular}
\end{table}

\section{Evaluation Prompts}
\noindent This section records the final GEPA candidates selected for E2H-Lichess and E2H-Codeforces, together with the hand-crafted AMC prompts, for \texttt{cot\_llm}, \texttt{debate}, \texttt{BoN}, and \texttt{self\_refine}.

\subsection{Chain-of-Thought LLM}
\subsubsection*{E2H-Lichess}
\begin{tabular}{@{}Lp{.76\textwidth}@{}}
\hline
\textbf{Prompt name} & \textbf{Prompt} \\
\hline
System prompt &
\prompttext{
Analyze the chess position step by step to find the correct mate-in-one move. Follow this systematic approach:

1. **Visualize the board**: Using the FEN, list all pieces of the side to move with their coordinates. Clearly note which piece is on each square.

2. **List all candidate checks**: Identify every piece that can deliver check to the opponent's king. For each candidate, strictly verify legality based on piece type:
   - **King**: can move exactly one square in any direction (orthogonal or diagonal). Do not consider moves that are two squares away.
   - **Pawn**: captures diagonally forward one square (direction depends on side: white moves up, black moves down). Forward pushes are not checks unless they capture an enemy piece? Actually a pawn giving check must attack the king diagonally; a forward push can give check if it moves to a square that attacks the king? No, pawns only attack diagonally. So only diagonal captures can deliver check. Ensure the capture target is occupied by an enemy piece.
   - **Knight**: moves in an L-shape (2+1). Ensure the target square is not occupied by a friendly piece.
   - **Sliding pieces (queen, rook, bishop)**: verify that the path is clear (no pieces between) and the target square is either empty or occupied by an enemy piece.

3. **Filter to legal checks**: Keep only candidates that are legal moves and actually give check. Double-check each candidate's movement rule---common mistake: confusing king moves with knight moves, or pawn captures with king captures.

4. **Verify checkmate** for each legal check:
   - The checking piece must be defended (the opponent's king cannot capture it safely, or if it can capture, that square is defended by another piece).
   - All possible king escape squares (including captures of the checking piece) must be either occupied by friendly pieces, blocked, or attacked by your pieces.
   - nfirm that no opponent piece can block or capture the checking piece to remove the check.

5. **Select the unique mate**: If only one legal check meets the criteria, it is the correct mate-in-one. If multiple, choose the one that eliminates all escapes.

6. **Output the move** as a JSON object with exactly:
   - "pgn": SAN in standard algebraic notation **without any annotation symbols** (no '+' for check, no '#' for checkmate). Use uppercase letters for pieces. Example: "Nxd6" not "Nxd6#".
   - "uci": UCI notation (e.g., "f5d6").
   Do not include any other text or commentary.

Ensure your reasoning is thorough and stepwise. Before outputting, re-verify that the UCI and SAN correspond to the same legal move and that the SAN has no annotations. Do not output anything except the JSON object.
} \\
\hline
\end{tabular}

\subsubsection*{E2H-Codeforces}
\begin{tabular}{@{}Lp{.76\textwidth}@{}}
\hline
\textbf{Prompt name} & \textbf{Prompt} \\
\hline
System prompt &
\prompttext{
Think step by step throh the competitive programming problem. Before writing code, analyze the problem, identify the algorithm, and consider edge cases. Then produce a Python 3.10 solution in a ```python fenced code block.
Important I/O requirements (STRICT): Read input using ONLY sys.stdin.readline(). Do NOT use sys.stdin.buffer.read().split(), sys.stdin.read(), or any other method that reads the entire input at once. For multiple test cases or large input, read lines one by one with a loop. Write output using sys.stdout.write() or print(), exactly matching the sample format. Always verify the exact format against the sample output.
Algorithm design: Verify small examples before coding. Handle empty input, single elements, duplicates, and boundaries. Validate array indices. Precompute prefix or suffix arrays for many queries. If using recursion, set sys.setrecursionlimit(10**6) and prefer iterative solutions. Check performance and test bitwise or combinatorial logic on small n.
Output format: The final answer must be a sing ```python code block containing the complete solution. Do not include text outside the code block. The code must be executable as a standalone script reading stdin and writing stdout.
Follow these guidelines to minimize runtime errors and ensure correct output.
} \\
\hline
\end{tabular}

\subsubsection*{AMC (Hand-Crafted)}
\begin{tabular}{@{}Lp{.76\textwidth}@{}}
\hline
\textbf{Prompt name} & \textbf{Prompt} \\
\hline
System prompt &
\prompttext{
Solve the high-school competition mathematics problem carefully.

Work step by step and give a clear, self-contained derivation. Before finalizing, check the arithmetic, logical conditions, and that the result answers exactly what was asked.

Do not use markdown fences or place intermediate results inside \boxed{...}. End with exactly one final-answer line:
Final answer: \boxed{...}
} \\
\hline
\end{tabular}

\subsection{Debate}
\subsubsection*{E2H-Lichess}
\begin{tabular}{@{}Lp{.76\textwidth}@{}}
\hline
\textbf{Prompt name} & \textbf{Prompt} \\
\hline
Generation system prompt &
\prompttext{
You are one agent in a multi-agent team.
Think step by step before giving your final answer.
Follow the output format specified in the task instructions.
} \\

Debate template &
\prompttext{
These are the solutions proposed by other agents:
{other_answers}

Carefully evaluate each proposed move. For each candidate, consider:
- Is it a legal move in the current position?
- Does it deliver checkmate immediately? (Remember: the puzzle is mate-in-one.)
- If multiple moves appear to checkmate, which one is correct based on the board?

Reason step-by-step to identify the unique winning move. Then, incorporating the insights (and correcting any errors) from the other agents, provide your final answer.

Your response must be a single valid JSON object with exactly two keys: "pgn" (a single SAN move, e.g., "Rxa2", with no check/mate symbols) and "uci" (the corresponding UCI string, e.g., "a1a2"). Output nothing else.

Original question:
{original_prompt}
} \\

Judge template &
\prompttext{
You are the judge. You will be given the original question and the final answers from multiple agents.
Synthesize the single best final answer.

Original question:
{original_prompt}

Agent answers:
{agent_answers}
} \\
\hline
\end{tabular}

\subsubsection*{E2H-Codeforces}
\begin{tabular}{@{}Lp{.76\textwidth}@{}}
\hline
\textbf{Prompt name} & \textbf{Prompt} \\
\hline
Generation system prompt &
\prompttext{
You are one agent in a multi-agent team solving Codeforces-style problems.
Analyze constraints, choose the right algorithm and data structures, then implement.
Return exactly one ```python fenced block with a complete runnable solution.
Avoid sys.stdin.buffer.* (unsupported in the test harness); sys.stdin.readline() or sys.stdin.read() are fine.
No text outside the code block.
} \\

Debate template &
\prompttext{
These are the solutions to the problem from other agents:
{other_answers}

Review each solution for algorithmic mistakes, missed edge cases, and wrong complexity.
Use the strongest ideas from other agents, fix errors, and provide your improved solution.

Original question:
{original_prompt}

Return a complete solution in one ```python block only.
} \\

Judge template &
\prompttext{
You are the judge. Pick or synthesize the single best final solution from the agent answers below.

Compare each candidate to the problem statement: correctness, constraints, edge cases, and whether it would pass hidden tests. If one agent answer is fully correct, output it verbatim in one ```python block. If several contain partial correct ideas, merge them into one working program.

Original question:
{original_prompt}

Agent answers:
{agent_answers}

Output only one ```python fenced block with the final solution.
} \\
\hline
\end{tabular}

\subsubsection*{AMC (Hand-Crafted)}
\begin{tabular}{@{}Lp{.76\textwidth}@{}}
\hline
\textbf{Prompt name} & \textbf{Prompt} \\
\hline
Generation system prompt &
\prompttext{
Solve and evaluate high-school competition mathematics carefully. Follow the specific orchestration role given in the user message: independent solver, revising debater, or final judge.

For any solution you produce, check arithmetic, logical conditions, and that the result answers exactly what was asked. Do not use markdown fences. End with exactly one final-answer line:
Final answer: \boxed{...}
} \\

Debate template &
\prompttext{
Act as a revising debate agent.

Other agents' latest solutions:
{other_answers}

Original task:
{original_prompt}

Critically compare their reasoning with your previous solution. Investigate disagreements instead of following the majority. Keep correct work, repair concrete errors, and account for any missed conditions.

Return a complete standalone revised solution. Do not mention the debate or refer to answers by agent number. End with exactly one final-answer line:
Final answer: \boxed{...}
} \\

Judge template &
\prompttext{
Act as the final judge.

Original task:
{original_prompt}

Final agent solutions:
{agent_answers}

Determine the correct answer by checking the agents' reasoning and calculations. Do not decide by majority vote. Resolve disagreements, repair errors, and synthesize a concise, self-contained final solution.

Do not mention the agents or the judging process. Do not use markdown fences. End with exactly one final-answer line:
Final answer: \boxed{...}
} \\
\hline
\end{tabular}

\subsection{Best-of-$N$}
\subsubsection*{E2H-Lichess}
\begin{tabular}{@{}Lp{.76\textwidth}@{}}\hline
\textbf{Prompt name} & \textbf{Prompt} \\ \hline
Generation system prompt & \tiny Think step by step through the problem before giving your final answer. Follow the output format specified in the task instructions. \\
Selector system prompt & \tiny Compare the candidates for the task and select the single best answer. Choose exactly one candidate index; do not rewrite or merge answers. \\
Selector template & \tiny You are selecting the best answer to the task below from \{num\_candidates\} candidates. Do not rewrite, merge, or synthesize a new answer. Original task: \{original\_prompt\}. Candidates: \{candidates\_block\}. Reply with exactly one final line (integer 1 to \{num\_candidates\}): Selected candidate: <number>. \\ \hline
\end{tabular}
\subsubsection*{E2H-Codeforces}
\begin{tabular}{@{}Lp{.76\textwidth}@{}}\hline
\textbf{Prompt name} & \textbf{Prompt} \\ \hline
Generation system prompt & \tiny Think step by step through the problem before writing code. Return exactly one ```python fenced block with a complete runnable solution. For reading input, use sys.stdin.readline() exclusively. Do NOT use sys.stdin.buffer.read(), sys.stdin.buffer.readline(), sys.stdin.read(), or any bulk read methods. Parse input line by line using readline() or iterating over sys.stdin. Use print() for output, ensuring each test case result is on its own line with no trailing spaces. No text outside the block. \\
Selector system prompt & \tiny Compare the candidates for the task and select the single best answer. Prefer the candidate with a complete ```python fenced solution. Choose exactly one candidate index; do not rewrite or merge answers. \\
Selector template & \tiny You are selecting the best answer to the task below from \{num\_candidates\} candidates. Do not rewrite, merge, or synthesize a new answer. Original task: \{original\_prompt\}. Candidates: \{candidates\_block\}. Reply with exactly one final line (integer 1 to \{num\_candidates\}): Selected candidate: <number>. \\ \hline
\end{tabular}

\subsubsection*{AMC (Hand-Crafted)}
\begin{tabular}{@{}Lp{.76\textwidth}@{}}
\hline
\textbf{Prompt name} & \textbf{Prompt} \\
\hline
Generation system prompt &
\prompttext{
Solve the high-school competition mathematics problem independently.

Give a clear, self-contained derivation. Check arithmetic, logical conditions, and that the result answers exactly what was asked. Do not assume access to other candidate solutions.

Do not use markdown fences. End with exactly one final-answer line:
Final answer: \boxed{...}
} \\

Selector system prompt &
\prompttext{
Act only as a selector. Choose the single candidate most likely to be mathematically correct.

Check each candidate's reasoning and final result against the original problem. Prefer correctness over confidence, length, or agreement with other candidates. Treat an unsupported answer or a formatting violation as evidence against a candidate.

Select an existing candidate by index. Do not solve the problem in your response, rewrite a candidate, or combine candidates.
} \\

Selector template &
\prompttext{
Select the best answer to the original task from the {num_candidates} candidates below.

Original task:
{original_prompt}

Candidates:
{candidates_block}

Compare the candidates for mathematical correctness, handling of conditions and edge cases, arithmetic accuracy, and compliance with the required final-answer format.

Return exactly one line containing an integer from 1 to {num_candidates}:
Selected candidate: <number>
} \\
\hline
\end{tabular}

\subsection{Self-Refine}
\subsubsection*{E2H-Lichess}
\begin{tabular}{@{}Lp{.76\textwidth}@{}}\hline
\textbf{Prompt name} & \textbf{Prompt} \\ \hline
Generation system prompt & \tiny Think step by step through the problem before giving your final answer. Follow the output format specified in the task instructions. \\
Feedback template & \tiny Review your answer to the task below. Give short, actionable feedback. Task: \{input\}. Your answer: \{current\_output\}. End with exactly one line: Is further refinement needed: yes/no \\
Refinement template & \tiny Improve your answer using the history below. Task: \{input\}. History: \{history\}. You are solving a mate-in-one chess puzzle. Carefully re-analyze the position: list all checks, verify every king escape, block, and capture, diagnose any previous error, and output only parseable JSON with exactly \{"pgn": single SAN without annotations, "uci": lowercase UCI\}. \\ \hline
\end{tabular}
\subsubsection*{E2H-Codeforces}
\begin{tabular}{@{}Lp{.76\textwidth}@{}}\hline
\textbf{Prompt name} & \textbf{Prompt} \\ \hline
Generation system prompt & \tiny Think step by step through the problem before giving your final answer. Return exactly one ```python fenced block with a complete runnable solution. CRITICAL: Use ONLY sys.stdin.readline() for reading input. Do NOT use sys.stdin.buffer.read(), sys.stdin.buffer.readline(), sys.stdin.read(), or any other buffer methods. These will cause a runtime\_error because the test harness uses StringIO. Use readline() for each line of input. Example: n = int(sys.stdin.readline()). You may also iterate with for line in sys.stdin: but each line must be read via readline. No text outside the block. \\
Feedback template & \tiny Review your answer to the task below. Provide short, actionable feedback focusing on correctness and compliance with the I/O contract. Task: \{input\}. Your answer: \{current\_output\}. Check sys.stdin.readline(), multiple test cases, exact output, edge cases, recursion safety, and off-by-one, type, or unused-variable errors. End with exactly one line: Is further refinement needed: yes/no \\
Refinement template & \tiny Improve your answer using the history below. Task: \{input\}. History: \{history\}. Output only one ```python fenced block with the final solution. \\ \hline
\end{tabular}

\subsubsection*{AMC (Hand-Crafted)}
\begin{tabular}{@{}Lp{.76\textwidth}@{}}
\hline
\textbf{Prompt name} & \textbf{Prompt} \\
\hline
Generation system prompt &
\prompttext{
Solve the high-school competition mathematics problem carefully and independently.

Give a clear, self-contained derivation. Check arithmetic, logical conditions, and that the result answers exactly what was asked.

Do not use markdown fences. End with exactly one final-answer line:
Final answer: \boxed{...}
} \\

Feedback template &
\prompttext{
Act only as a critic of the proposed solution.

Task:
{input}

Proposed solution:
{current_output}

Check the reasoning, arithmetic, interpretation of the problem, and final-answer format. Identify concrete errors, unsupported steps, or missing cases. If it is correct, say so briefly. Do not rewrite the solution and do not produce a new boxed answer.

The answer is format-valid only if it ends with exactly one line of the form:
Final answer: \boxed{...}

End with exactly one of these lines:
Is further refinement needed: yes
Is further refinement needed: no
} \\

Refinement template &
\prompttext{
Produce the improved final response using the task and refinement history below.

Task:
{input}

Refinement history:
{history}

Correct the identified issues and return a standalone solution. Do not discuss the refinement process or refer to the critic. Recheck the final result.

Do not use markdown fences. End with exactly one final-answer line:
Final answer: \boxed{...}
} \\
\hline
\end{tabular}

\section{Model Configurations}
\begin{table}[H]
\centering
\small
\setlength{\tabcolsep}{3pt}
\renewcommand{\arraystretch}{1.2}
\begin{threeparttable}
\caption{Backbone configurations.
All methods within an experiment share one backbone and decoding configuration.}
\label{tab:model-configurations}
\begin{tabularx}{\textwidth}{@{}>{\raggedright\arraybackslash}X l l >{\raggedright\arraybackslash}X@{}}
\toprule
model & parameters & quantization & hardware \\
\midrule
\path{deepseek-ai/DeepSeek-V4-Flash} & 284B (13B active) & FP4 + FP8 mixed & 4 $\times$ H200 141\,GB \\
\path{google/gemma-4-26B-A4B-it} & 25.2B (3.8B active) & BF16 & 1 $\times$ H200 141\,GB \\
\path{zai-org/GLM-4.7-Flash} & 30B (3B active) & BF16 & 1 $\times$ H200 141\,GB \\
\path{Qwen/Qwen3.6-35B-A3B-FP8} & 35B (3B active) & FP8 & 2 $\times$ A100 80\,GB \\
\path{nvidia/gpt-oss-puzzle-88B} & 88B & MXFP4 weights; FP8 KV cache & 1 $\times$ H200 141\,GB \\
\bottomrule
\end{tabularx}
\end{threeparttable}
\end{table}

\section{Continuous Difficulty Profiles}

The figure below shows descriptive absolute-accuracy trajectories across ten within-benchmark difficulty bins.
Accuracy generally decreases with difficulty, although the trajectories are not monotonic in every benchmark.
These binned estimates provide descriptive context for the mixed-model analysis and should not be interpreted as fitted trends or inferential comparisons.

\begin{figure}[H]
\centering
\includegraphics[width=0.75\textwidth]{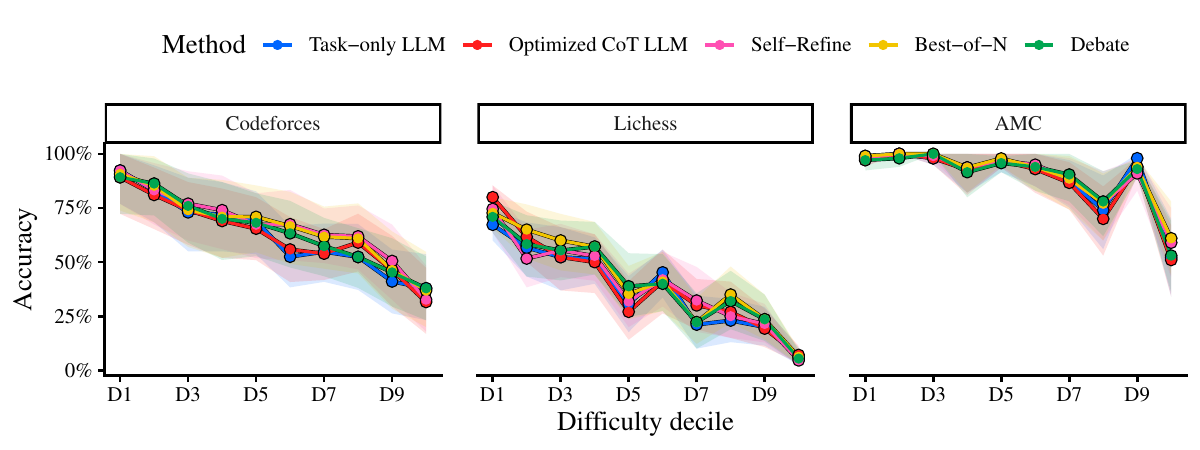}
\caption{Absolute accuracy over ten within-benchmark bins of the continuous difficulty percentile.
Each point aggregates the five evaluated models over the held-out items in that bin; lines connect descriptive bin estimates and do not represent fitted models or uncertainty intervals.}
\label{fig:continuous-difficulty}
\end{figure}

\section{Statistical Tests}

Table~\ref{tab:mixed-model-fits} summarizes the fit and diagnostics of Models~$M_0$--$M_3$.
Table~\ref{tab:mixed-model-comparisons} reports the likelihood-ratio comparisons, and Tables~\ref{tab:average-method-contrasts} and~\ref{tab:difficulty-tests} report the planned contrasts.

\begin{table}[H]
\centering
\footnotesize
\setlength{\tabcolsep}{3pt}
\caption{Mixed-logistic-model fit statistics and diagnostics.
Each model is fitted to 4{,}000 observations comprising 200 items and 1{,}000 item-by-LLM groups.
All models include random intercepts for item and item-by-LLM, and all converged without singular fits.
Models~$M_0$--$M_2$ are confirmatory, whereas $M_3$ is exploratory.}
\label{tab:mixed-model-fits}
\begin{tabular}{@{}lllrr@{}}
\toprule
benchmark & model & fixed-effects structure & log likelihood & AIC \\
\midrule
Codeforces & $M_0$ & difficulty + LLM & $-1300.429$ & 2616.858 \\
 & $M_1$ & $M_0$ + method & $-1291.119$ & 2604.237 \\
 & $M_2$ & $M_1$ + method $\times$ difficulty & $-1290.382$ & 2608.764 \\
 & $M_3$ & $M_1$ + method $\times$ LLM & $-1264.210$ & 2574.419 \\
\midrule
Lichess & $M_0$ & difficulty + LLM & $-1298.308$ & 2612.616 \\
 & $M_1$ & $M_0$ + method & $-1296.137$ & 2614.274 \\
 & $M_2$ & $M_1$ + method $\times$ difficulty & $-1296.021$ & 2620.041 \\
 & $M_3$ & $M_1$ + method $\times$ LLM & $-1280.797$ & 2607.594 \\
\midrule
AMC & $M_0$ & difficulty + LLM & $-563.940$ & 1143.880 \\
 & $M_1$ & $M_0$ + method & $-557.058$ & 1136.117 \\
 & $M_2$ & $M_1$ + method $\times$ difficulty & $-556.710$ & 1141.421 \\
 & $M_3$ & $M_1$ + method $\times$ LLM & $-513.220$ & 1072.439 \\
\bottomrule
\end{tabular}
\end{table}

Models were fitted in R~4.5.1 using \texttt{lme4}~1.1-37, a binomial likelihood with a logit link, and the \texttt{bobyqa} optimizer with \texttt{maxfun = 200000}.

\begin{table}[H]
\centering
\footnotesize
\setlength{\tabcolsep}{4pt}
\caption{Likelihood-ratio comparisons among the mixed logistic models.
The $M_1$--$M_3$ comparisons are exploratory; all others are confirmatory.}
\label{tab:mixed-model-comparisons}
\begin{tabular}{@{}lllrrr@{}}
\toprule
benchmark & comparison & tested effect & $\chi^2$ & df & $p$ \\
\midrule
Codeforces & $M_0$ vs.\ $M_1$ & average method effect & 18.6210 & 3 & .0003 \\
 & $M_1$ vs.\ $M_2$ & method $\times$ difficulty & 1.4735 & 3 & .6884 \\
 & $M_1$ vs.\ $M_3$ & method $\times$ LLM & 53.8179 & 12 & $<.0001$ \\
\midrule
Lichess & $M_0$ vs.\ $M_1$ & average method effect & 4.3426 & 3 & .2268 \\
 & $M_1$ vs.\ $M_2$ & method $\times$ difficulty & .2325 & 3 & .9722 \\
 & $M_1$ vs.\ $M_3$ & method $\times$ LLM & 30.6799 & 12 & .0022 \\
\midrule
AMC & $M_0$ vs.\ $M_1$ & average method effect & 13.7635 & 3 & .0032 \\
 & $M_1$ vs.\ $M_2$ & method $\times$ difficulty & .6960 & 3 & .8741 \\
 & $M_1$ vs.\ $M_3$ & method $\times$ LLM & 87.6773 & 12 & $<.0001$ \\
\bottomrule
\end{tabular}
\end{table}

\begin{table}[H]
\centering
\footnotesize
\setlength{\tabcolsep}{4pt}
\caption{Planned average orchestration-versus-optimized-CoT contrasts from $M_1$.
Intervals are nominal 95\% confidence intervals, and $p_{\mathrm{Holm}}$ adjusts the three contrasts within each benchmark.}
\label{tab:average-method-contrasts}
\begin{tabular}{@{}llrrrr@{}}
\toprule
benchmark & orchestration & odds ratio & 95\% CI & $p$ & $p_{\mathrm{Holm}}$ \\
\midrule
Codeforces & Self-Refine & 1.8999 & [1.3645, 2.6455] & .0001 & .0004 \\
 & BoN & 1.7916 & [1.2886, 2.4910] & .0005 & .0010 \\
 & Debate & 1.3476 & [.9753, 1.8621] & .0706 & .0706 \\
\midrule
Lichess & Self-Refine & .9597 & [.6939, 1.3273] & .8038 & .9158 \\
 & BoN & 1.3103 & [.9489, 1.8092] & .1007 & .3021 \\
 & Debate & 1.1302 & [.8181, 1.5614] & .4579 & .9158 \\
\midrule
AMC & Self-Refine & 1.9905 & [1.1642, 3.4031] & .0119 & .0238 \\
 & BoN & 2.6802 & [1.5341, 4.6824] & .0005 & .0016 \\
 & Debate & 1.5175 & [.9023, 2.5522] & .1158 & .1158 \\
\bottomrule
\end{tabular}
\end{table}

\begin{table}[H]
\centering
\footnotesize
\setlength{\tabcolsep}{3pt}
\caption{Planned orchestration-versus-optimized-CoT difficulty-slope contrasts from $M_2$.
Slope ratios are changes in the conditional orchestration-versus-optimized-CoT odds ratio per one-standard-deviation increase in continuous difficulty.
Contrast intervals are nominal 95\% intervals, and $p_{\mathrm{Holm}}$ adjusts the three contrasts within each benchmark.}
\label{tab:difficulty-tests}
\begin{tabular}{@{}llrrrr@{}}
\toprule
benchmark & orchestration & slope ratio & 95\% CI & $p$ & $p_{\mathrm{Holm}}$ \\
\midrule
Codeforces & Self-Refine & .8205 & [.5664, 1.1887] & .2955 & .8866 \\
 & BoN & .8842 & [.6131, 1.2754] & .5103 & .8866 \\
 & Debate & .8260 & [.5754, 1.1859] & .3003 & .8866 \\
\midrule
Lichess & Self-Refine & 1.0420 & [.7413, 1.4647] & .8129 & 1.0000 \\
 & BoN & .9741 & [.6932, 1.3688] & .8796 & 1.0000 \\
 & Debate & 1.0454 & [.7447, 1.4677] & .7973 & 1.0000 \\
\midrule
AMC & Self-Refine & 1.0320 & [.5673, 1.8774] & .9178 & 1.0000 \\
 & BoN & .9148 & [.4836, 1.7303] & .7841 & 1.0000 \\
 & Debate & 1.1864 & [.6704, 2.0995] & .5572 & 1.0000 \\
\bottomrule
\end{tabular}
\end{table}

\section{Complete Model--Method Results}

Tables~\ref{tab:complete-results-codeforces}--\ref{tab:complete-results-amc} report all 75 model--method cells.
Brackets are descriptive 95\% whole-item bootstrap intervals.
Weighted-token totals include all attempts, including failed attempts.
The final column is the ratio of weighted tokens per correct answer to the corresponding task-only cell.

\begin{table}[H]
\centering
\tiny
\setlength{\tabcolsep}{2pt}
\caption{Complete Codeforces results.}
\label{tab:complete-results-codeforces}
\resizebox{\textwidth}{!}{%
\begin{tabular}{@{}lllrr@{}}
\toprule
model & method & accuracy [95\% CI] & weighted tokens/correct [95\% CI] & relative to task-only [95\% CI] \\
\midrule
DeepSeek V4 Flash & Task-only & 60.0\% [53.0, 67.0] & 63,448 [50,676, 78,831] & 1.00x [1.00, 1.00] \\
DeepSeek V4 Flash & CoT & 70.0\% [63.5, 76.5] & 56,700 [46,442, 68,986] & 0.89x [0.78, 1.03] \\
DeepSeek V4 Flash & Self-Refine & 72.5\% [66.0, 78.5] & 109,027 [89,055, 133,050] & 1.72x [1.51, 1.95] \\
DeepSeek V4 Flash & Best-of-$N$ & 66.0\% [59.5, 72.5] & 217,353 [176,364, 267,289] & 3.43x [3.03, 3.89] \\
DeepSeek V4 Flash & Debate & 69.5\% [63.0, 76.0] & 210,583 [172,517, 256,962] & 3.32x [2.91, 3.79] \\
Gemma 4 & Task-only & 74.5\% [68.5, 80.5] & 57,142 [50,853, 64,667] & 1.00x [1.00, 1.00] \\
Gemma 4 & CoT & 69.5\% [63.0, 76.0] & 88,645 [72,548, 109,232] & 1.55x [1.31, 1.84] \\
Gemma 4 & Self-Refine & 74.5\% [68.5, 80.5] & 181,073 [155,508, 215,594] & 3.17x [2.86, 3.55] \\
Gemma 4 & Best-of-$N$ & 72.5\% [66.0, 78.5] & 255,684 [229,094, 288,941] & 4.47x [4.10, 4.90] \\
Gemma 4 & Debate & 74.0\% [67.5, 80.0] & 294,238 [258,461, 337,743] & 5.15x [4.78, 5.57] \\
GLM 4.7 Flash & Task-only & 49.0\% [42.0, 56.0] & 261,377 [207,293, 332,276] & 1.00x [1.00, 1.00] \\
GLM 4.7 Flash & CoT & 38.5\% [32.0, 45.5] & 256,298 [199,960, 334,060] & 0.98x [0.79, 1.22] \\
GLM 4.7 Flash & Self-Refine & 50.0\% [43.0, 57.0] & 498,143 [396,445, 632,264] & 1.91x [1.58, 2.30] \\
GLM 4.7 Flash & Best-of-$N$ & 56.5\% [49.5, 63.0] & 801,885 [656,827, 990,135] & 3.07x [2.65, 3.55] \\
GLM 4.7 Flash & Debate & 37.0\% [30.0, 43.5] & 850,633 [671,959, 1,102,147] & 3.25x [2.66, 4.03] \\
Qwen 3.6 & Task-only & 66.5\% [60.0, 73.0] & 182,593 [156,780, 213,485] & 1.00x [1.00, 1.00] \\
Qwen 3.6 & CoT & 65.5\% [58.5, 72.0] & 154,054 [129,388, 184,543] & 0.84x [0.74, 0.96] \\
Qwen 3.6 & Self-Refine & 66.0\% [59.5, 72.5] & 257,994 [218,567, 305,966] & 1.41x [1.27, 1.57] \\
Qwen 3.6 & Best-of-$N$ & 67.5\% [61.0, 74.0] & 516,287 [449,732, 595,899] & 2.83x [2.58, 3.09] \\
Qwen 3.6 & Debate & 68.0\% [61.5, 74.5] & 474,897 [408,839, 557,005] & 2.60x [2.34, 2.89] \\
GPT-OSS Puzzle & Task-only & 58.0\% [51.0, 65.0] & 37,024 [29,370, 46,669] & 1.00x [1.00, 1.00] \\
GPT-OSS Puzzle & CoT & 64.0\% [57.5, 70.5] & 32,856 [27,015, 40,090] & 0.89x [0.76, 1.04] \\
GPT-OSS Puzzle & Self-Refine & 67.5\% [61.0, 74.0] & 72,802 [58,104, 90,667] & 1.97x [1.66, 2.32] \\
GPT-OSS Puzzle & Best-of-$N$ & 66.0\% [59.5, 72.5] & 123,230 [102,486, 148,285] & 3.33x [2.91, 3.81] \\
GPT-OSS Puzzle & Debate & 70.0\% [63.5, 76.5] & 130,888 [110,057, 155,927] & 3.54x [3.04, 4.10] \\
\bottomrule
\end{tabular}}
\end{table}

\begin{table}[H]
\centering
\tiny
\setlength{\tabcolsep}{2pt}
\caption{Complete Lichess results.}
\label{tab:complete-results-lichess}
\resizebox{\textwidth}{!}{%
\begin{tabular}{@{}lllrr@{}}
\toprule
model & method & accuracy [95\% CI] & weighted tokens/correct [95\% CI] & relative to task-only [95\% CI] \\
\midrule
DeepSeek V4 Flash & Task-only & 42.0\% [35.0, 49.0] & 225,183 [185,799, 278,523] & 1.00x [1.00, 1.00] \\
DeepSeek V4 Flash & CoT & 50.5\% [43.5, 57.5] & 185,510 [156,083, 223,735] & 0.82x [0.70, 0.97] \\
DeepSeek V4 Flash & Self-Refine & 47.5\% [40.5, 54.5] & 633,122 [518,788, 779,534] & 2.81x [2.34, 3.37] \\
DeepSeek V4 Flash & Best-of-$N$ & 46.5\% [40.0, 53.5] & 774,616 [648,892, 941,308] & 3.44x [2.92, 4.01] \\
DeepSeek V4 Flash & Debate & 44.0\% [37.0, 50.5] & 889,996 [740,873, 1,093,724] & 3.95x [3.33, 4.67] \\
Gemma 4 & Task-only & 33.5\% [27.0, 40.0] & 467,421 [344,487, 639,927] & 1.00x [1.00, 1.00] \\
Gemma 4 & CoT & 31.0\% [24.5, 37.5] & 843,592 [633,614, 1,147,410] & 1.80x [1.34, 2.46] \\
Gemma 4 & Self-Refine & 34.5\% [28.0, 41.0] & 662,449 [501,243, 899,207] & 1.42x [1.05, 1.91] \\
Gemma 4 & Best-of-$N$ & 43.0\% [36.0, 50.0] & 1,015,592 [798,676, 1,303,091] & 2.17x [1.67, 2.82] \\
Gemma 4 & Debate & 37.0\% [30.0, 43.5] & 1,328,866 [1,025,086, 1,768,367] & 2.84x [2.21, 3.66] \\
GLM 4.7 Flash & Task-only & 5.0\% [2.0, 8.0] & 2,653,007 [1,598,669, 6,081,422] & 1.00x [1.00, 1.00] \\
GLM 4.7 Flash & CoT & 3.0\% [1.0, 5.5] & 4,432,704 [2,343,823, 13,686,754] & 1.67x [0.61, 5.85] \\
GLM 4.7 Flash & Self-Refine & 3.0\% [1.0, 5.5] & 15,599,188 [8,383,170, 47,584,239] & 5.88x [2.20, 20.07] \\
GLM 4.7 Flash & Best-of-$N$ & 3.5\% [1.0, 6.0] & 13,143,909 [7,339,854, 44,895,079] & 4.95x [2.02, 14.08] \\
GLM 4.7 Flash & Debate & 3.5\% [1.0, 6.0] & 15,336,457 [8,698,859, 52,364,064] & 5.78x [2.21, 18.17] \\
Qwen 3.6 & Task-only & 20.0\% [14.5, 25.5] & 185,344 [140,968, 263,236] & 1.00x [1.00, 1.00] \\
Qwen 3.6 & CoT & 27.0\% [21.0, 33.5] & 161,080 [126,601, 213,164] & 0.87x [0.65, 1.14] \\
Qwen 3.6 & Self-Refine & 24.0\% [18.0, 30.0] & 286,497 [220,728, 389,582] & 1.55x [1.19, 1.95] \\
Qwen 3.6 & Best-of-$N$ & 22.5\% [16.5, 28.5] & 539,536 [417,003, 737,403] & 2.91x [2.21, 3.80] \\
Qwen 3.6 & Debate & 24.0\% [18.0, 30.0] & 554,802 [436,384, 744,392] & 2.99x [2.31, 3.79] \\
GPT-OSS Puzzle & Task-only & 47.0\% [40.0, 54.0] & 124,511 [102,928, 153,549] & 1.00x [1.00, 1.00] \\
GPT-OSS Puzzle & CoT & 43.5\% [36.5, 50.5] & 141,864 [116,627, 176,844] & 1.14x [1.02, 1.28] \\
GPT-OSS Puzzle & Self-Refine & 44.5\% [37.5, 51.5] & 447,084 [363,439, 560,338] & 3.59x [3.24, 3.98] \\
GPT-OSS Puzzle & Best-of-$N$ & 49.5\% [42.5, 56.5] & 392,882 [329,154, 479,693] & 3.16x [2.83, 3.48] \\
GPT-OSS Puzzle & Debate & 51.0\% [44.0, 58.0] & 489,426 [409,269, 594,538] & 3.93x [3.53, 4.35] \\
\bottomrule
\end{tabular}}
\end{table}

\begin{table}[H]
\centering
\tiny
\setlength{\tabcolsep}{2pt}
\caption{Complete AMC results.}
\label{tab:complete-results-amc}
\resizebox{\textwidth}{!}{%
\begin{tabular}{@{}lllrr@{}}
\toprule
model & method & accuracy [95\% CI] & weighted tokens/correct [95\% CI] & relative to task-only [95\% CI] \\
\midrule
DeepSeek V4 Flash & Task-only & 93.0\% [89.5, 96.0] & 51,139 [38,294, 66,175] & 1.00x [1.00, 1.00] \\
DeepSeek V4 Flash & CoT & 92.0\% [88.0, 95.5] & 61,445 [46,605, 79,223] & 1.20x [1.09, 1.34] \\
DeepSeek V4 Flash & Self-Refine & 95.0\% [92.0, 98.0] & 105,142 [77,038, 139,407] & 2.06x [1.74, 2.38] \\
DeepSeek V4 Flash & Best-of-$N$ & 94.5\% [91.0, 97.5] & 185,347 [144,499, 232,293] & 3.62x [3.26, 4.03] \\
DeepSeek V4 Flash & Debate & 93.5\% [90.0, 96.5] & 179,000 [133,689, 233,730] & 3.50x [3.13, 3.89] \\
Gemma 4 & Task-only & 90.5\% [86.5, 94.5] & 72,296 [55,469, 92,275] & 1.00x [1.00, 1.00] \\
Gemma 4 & CoT & 81.5\% [76.0, 86.5] & 126,806 [95,346, 164,949] & 1.75x [1.37, 2.26] \\
Gemma 4 & Self-Refine & 88.5\% [84.0, 92.5] & 169,645 [129,262, 217,756] & 2.35x [1.83, 3.00] \\
Gemma 4 & Best-of-$N$ & 88.0\% [83.5, 92.5] & 358,775 [291,831, 434,989] & 4.96x [4.13, 6.04] \\
Gemma 4 & Debate & 92.5\% [88.5, 96.0] & 359,564 [287,479, 442,465] & 4.97x [4.14, 6.03] \\
GLM 4.7 Flash & Task-only & 88.0\% [83.5, 92.0] & 80,188 [62,137, 101,146] & 1.00x [1.00, 1.00] \\
GLM 4.7 Flash & CoT & 87.5\% [83.0, 92.0] & 85,498 [67,945, 106,086] & 1.07x [0.95, 1.20] \\
GLM 4.7 Flash & Self-Refine & 88.5\% [84.0, 92.5] & 177,634 [137,064, 227,232] & 2.22x [1.91, 2.57] \\
GLM 4.7 Flash & Best-of-$N$ & 90.0\% [85.5, 94.0] & 288,924 [234,782, 352,831] & 3.60x [3.27, 4.02] \\
GLM 4.7 Flash & Debate & 78.0\% [72.5, 83.5] & 288,724 [240,757, 345,951] & 3.60x [3.09, 4.23] \\
Qwen 3.6 & Task-only & 89.5\% [85.0, 93.5] & 106,985 [92,818, 123,074] & 1.00x [1.00, 1.00] \\
Qwen 3.6 & CoT & 92.5\% [88.5, 96.0] & 102,565 [90,403, 115,830] & 0.96x [0.90, 1.02] \\
Qwen 3.6 & Self-Refine & 91.5\% [87.5, 95.0] & 194,616 [163,538, 230,438] & 1.82x [1.67, 1.98] \\
Qwen 3.6 & Best-of-$N$ & 93.0\% [89.0, 96.5] & 323,402 [286,603, 364,327] & 3.02x [2.85, 3.20] \\
Qwen 3.6 & Debate & 93.0\% [89.0, 96.5] & 279,302 [244,831, 318,335] & 2.61x [2.46, 2.78] \\
GPT-OSS Puzzle & Task-only & 85.5\% [80.5, 90.0] & 26,027 [20,330, 33,009] & 1.00x [1.00, 1.00] \\
GPT-OSS Puzzle & CoT & 86.0\% [81.0, 90.5] & 23,694 [18,985, 29,205] & 0.91x [0.81, 1.03] \\
GPT-OSS Puzzle & Self-Refine & 85.5\% [80.5, 90.0] & 61,502 [45,498, 80,922] & 2.36x [2.07, 2.65] \\
GPT-OSS Puzzle & Best-of-$N$ & 87.0\% [82.0, 91.5] & 99,146 [79,578, 122,713] & 3.81x [3.45, 4.22] \\
GPT-OSS Puzzle & Debate & 88.5\% [84.0, 92.5] & 105,160 [84,278, 130,380] & 4.04x [3.70, 4.42] \\
\bottomrule
\end{tabular}}
\end{table}

\section{Model-Specific Paired Contrasts}

Table~\ref{tab:model-specific-contrasts} reports the complete exploratory model-specific paired effects relative to optimized CoT.
Each row contains 200 paired items.
The contrasts are descriptive; no model-specific hypothesis-test family was defined.
Wins and losses count items on which the orchestration was correct and CoT was incorrect, or vice versa.

\begin{table}[H]
\centering
\scriptsize
\setlength{\tabcolsep}{3pt}
\caption{Complete exploratory model-specific paired contrasts relative to optimized CoT.
The difference is in percentage points.}
\label{tab:model-specific-contrasts}
\resizebox{\textwidth}{!}{%
\begin{tabular}{@{}lllrrrrrr@{}}
\toprule
benchmark & model & orchestration & CoT acc. & orchestration acc. & difference & wins & losses & ties \\
\midrule
Codeforces & DeepSeek V4 Flash & Best-of-$N$ & 70.0\% & 66.0\% & \textcolor{red}{-4.0} & 12 & 20 & 168 \\
Codeforces & DeepSeek V4 Flash & Debate & 70.0\% & 69.5\% & \textcolor{red}{-0.5} & 10 & 11 & 179 \\
Codeforces & DeepSeek V4 Flash & Self-Refine & 70.0\% & 72.5\% & \textcolor{green!50!black}{+2.5} & 14 & 9 & 177 \\
Codeforces & Gemma 4 & Best-of-$N$ & 69.5\% & 72.5\% & \textcolor{green!50!black}{+3.0} & 12 & 6 & 182 \\
Codeforces & Gemma 4 & Debate & 69.5\% & 74.0\% & \textcolor{green!50!black}{+4.5} & 14 & 5 & 181 \\
Codeforces & Gemma 4 & Self-Refine & 69.5\% & 74.5\% & \textcolor{green!50!black}{+5.0} & 13 & 3 & 184 \\
Codeforces & GLM 4.7 Flash & Best-of-$N$ & 38.5\% & 56.5\% & \textcolor{green!50!black}{+18.0} & 45 & 9 & 146 \\
Codeforces & GLM 4.7 Flash & Debate & 38.5\% & 37.0\% & \textcolor{red}{-1.5} & 21 & 24 & 155 \\
Codeforces & GLM 4.7 Flash & Self-Refine & 38.5\% & 50.0\% & \textcolor{green!50!black}{+11.5} & 43 & 20 & 137 \\
Codeforces & Qwen 3.6 & Best-of-$N$ & 65.5\% & 67.5\% & \textcolor{green!50!black}{+2.0} & 14 & 10 & 176 \\
Codeforces & Qwen 3.6 & Debate & 65.5\% & 68.0\% & \textcolor{green!50!black}{+2.5} & 19 & 14 & 167 \\
Codeforces & Qwen 3.6 & Self-Refine & 65.5\% & 66.0\% & \textcolor{green!50!black}{+0.5} & 18 & 17 & 165 \\
Codeforces & GPT-OSS Puzzle & Best-of-$N$ & 64.0\% & 66.0\% & \textcolor{green!50!black}{+2.0} & 15 & 11 & 174 \\
Codeforces & GPT-OSS Puzzle & Debate & 64.0\% & 70.0\% & \textcolor{green!50!black}{+6.0} & 22 & 10 & 168 \\
Codeforces & GPT-OSS Puzzle & Self-Refine & 64.0\% & 67.5\% & \textcolor{green!50!black}{+3.5} & 15 & 8 & 177 \\
\midrule
Lichess & DeepSeek V4 Flash & Best-of-$N$ & 50.5\% & 46.5\% & \textcolor{red}{-4.0} & 15 & 23 & 162 \\
Lichess & DeepSeek V4 Flash & Debate & 50.5\% & 44.0\% & \textcolor{red}{-6.5} & 15 & 28 & 157 \\
Lichess & DeepSeek V4 Flash & Self-Refine & 50.5\% & 47.5\% & \textcolor{red}{-3.0} & 26 & 32 & 142 \\
Lichess & Gemma 4 & Best-of-$N$ & 31.0\% & 43.0\% & \textcolor{green!50!black}{+12.0} & 32 & 8 & 160 \\
Lichess & Gemma 4 & Debate & 31.0\% & 37.0\% & \textcolor{green!50!black}{+6.0} & 25 & 13 & 162 \\
Lichess & Gemma 4 & Self-Refine & 31.0\% & 34.5\% & \textcolor{green!50!black}{+3.5} & 21 & 14 & 165 \\
Lichess & GLM 4.7 Flash & Best-of-$N$ & 3.0\% & 3.5\% & \textcolor{green!50!black}{+0.5} & 7 & 6 & 187 \\
Lichess & GLM 4.7 Flash & Debate & 3.0\% & 3.5\% & \textcolor{green!50!black}{+0.5} & 7 & 6 & 187 \\
Lichess & GLM 4.7 Flash & Self-Refine & 3.0\% & 3.0\% & +0.0 & 5 & 5 & 190 \\
Lichess & Qwen 3.6 & Best-of-$N$ & 27.0\% & 22.5\% & \textcolor{red}{-4.5} & 15 & 24 & 161 \\
Lichess & Qwen 3.6 & Debate & 27.0\% & 24.0\% & \textcolor{red}{-3.0} & 13 & 19 & 168 \\
Lichess & Qwen 3.6 & Self-Refine & 27.0\% & 24.0\% & \textcolor{red}{-3.0} & 11 & 17 & 172 \\
Lichess & GPT-OSS Puzzle & Best-of-$N$ & 43.5\% & 49.5\% & \textcolor{green!50!black}{+6.0} & 18 & 6 & 176 \\
Lichess & GPT-OSS Puzzle & Debate & 43.5\% & 51.0\% & \textcolor{green!50!black}{+7.5} & 21 & 6 & 173 \\
Lichess & GPT-OSS Puzzle & Self-Refine & 43.5\% & 44.5\% & \textcolor{green!50!black}{+1.0} & 13 & 11 & 176 \\
\midrule
AMC & DeepSeek V4 Flash & Best-of-$N$ & 92.0\% & 94.5\% & \textcolor{green!50!black}{+2.5} & 5 & 0 & 195 \\
AMC & DeepSeek V4 Flash & Debate & 92.0\% & 93.5\% & \textcolor{green!50!black}{+1.5} & 4 & 1 & 195 \\
AMC & DeepSeek V4 Flash & Self-Refine & 92.0\% & 95.0\% & \textcolor{green!50!black}{+3.0} & 6 & 0 & 194 \\
AMC & Gemma 4 & Best-of-$N$ & 81.5\% & 88.0\% & \textcolor{green!50!black}{+6.5} & 16 & 3 & 181 \\
AMC & Gemma 4 & Debate & 81.5\% & 92.5\% & \textcolor{green!50!black}{+11.0} & 23 & 1 & 176 \\
AMC & Gemma 4 & Self-Refine & 81.5\% & 88.5\% & \textcolor{green!50!black}{+7.0} & 17 & 3 & 180 \\
AMC & GLM 4.7 Flash & Best-of-$N$ & 87.5\% & 90.0\% & \textcolor{green!50!black}{+2.5} & 8 & 3 & 189 \\
AMC & GLM 4.7 Flash & Debate & 87.5\% & 78.0\% & \textcolor{red}{-9.5} & 5 & 24 & 171 \\
AMC & GLM 4.7 Flash & Self-Refine & 87.5\% & 88.5\% & \textcolor{green!50!black}{+1.0} & 7 & 5 & 188 \\
AMC & Qwen 3.6 & Best-of-$N$ & 92.5\% & 93.0\% & \textcolor{green!50!black}{+0.5} & 2 & 1 & 197 \\
AMC & Qwen 3.6 & Debate & 92.5\% & 93.0\% & \textcolor{green!50!black}{+0.5} & 2 & 1 & 197 \\
AMC & Qwen 3.6 & Self-Refine & 92.5\% & 91.5\% & \textcolor{red}{-1.0} & 2 & 4 & 194 \\
AMC & GPT-OSS Puzzle & Best-of-$N$ & 86.0\% & 87.0\% & \textcolor{green!50!black}{+1.0} & 7 & 5 & 188 \\
AMC & GPT-OSS Puzzle & Debate & 86.0\% & 88.5\% & \textcolor{green!50!black}{+2.5} & 8 & 3 & 189 \\
AMC & GPT-OSS Puzzle & Self-Refine & 86.0\% & 85.5\% & \textcolor{red}{-0.5} & 7 & 8 & 185 \\
\bottomrule
\end{tabular}}
\end{table}

\section{Recorded Failure Counts}

Table~\ref{tab:failure-counts} reports every model--method cell with at least one recorded or conservatively inferred technical failure.
Cells omitted from the table had no such failures.
An output-limit failure is an explicit \texttt{MaxTokensExceeded} outcome, a score-zero evaluation whose aggregate completion count reached \(120{,}000\) times the number of API calls, or a recorded \texttt{ContextWindowExceeded} outcome.
The percentage is relative to all score-zero evaluations in the cell.
Actual wrong items are score-zero evaluations remaining after recorded errors and inferred output-limit failures are excluded.

\begin{table}[H]
\centering
\scriptsize
\setlength{\tabcolsep}{3pt}
\caption{Recorded or conservatively inferred technical-failure outcomes.
Each model--method cell contains 200 evaluations.}
\label{tab:failure-counts}
\resizebox{\textwidth}{!}{%
\begin{tabular}{@{}lllrrrr@{}}
\toprule
benchmark & model & method & output-limit failures & \% of incorrect & other failures & actual wrong items \\
\midrule
Codeforces & Gemma 4 & CoT & 5 & 8.2\% & 0 & 56 \\
Codeforces & GLM 4.7 Flash & Task-only & 11 & 10.8\% & 0 & 91 \\
Codeforces & GLM 4.7 Flash & CoT & 4 & 3.3\% & 0 & 119 \\
Codeforces & Qwen 3.6 & Task-only & 1 & 1.5\% & 0 & 66 \\
Codeforces & Qwen 3.6 & CoT & 2 & 2.9\% & 0 & 67 \\
Lichess & Gemma 4 & Task-only & 49 & 36.8\% & 0 & 84 \\
Lichess & Gemma 4 & CoT & 97 & 70.3\% & 0 & 41 \\
Lichess & Gemma 4 & Debate & 12 & 9.5\% & 0 & 114 \\
Lichess & GLM 4.7 Flash & Task-only & 8 & 4.2\% & 0 & 182 \\
Lichess & GLM 4.7 Flash & CoT & 6 & 3.1\% & 0 & 188 \\
AMC & DeepSeek V4 Flash & Task-only & 3 & 21.4\% & 0 & 11 \\
AMC & DeepSeek V4 Flash & CoT & 6 & 37.5\% & 0 & 10 \\
AMC & Gemma 4 & Task-only & 11 & 57.9\% & 0 & 8 \\
AMC & Gemma 4 & CoT & 29 & 78.4\% & 0 & 8 \\
AMC & GLM 4.7 Flash & Task-only & 5 & 20.8\% & 0 & 19 \\
AMC & GLM 4.7 Flash & CoT & 2 & 8.0\% & 0 & 23 \\
AMC & GLM 4.7 Flash & Best-of-$N$ & 1 & 5.0\% & 0 & 19 \\
\bottomrule
\end{tabular}}
\end{table}

\end{document}